\documentclass[[12pt]{article}
\usepackage{graphicx}

\usepackage{formatfig}
\usepackage{graphicx}
\usepackage{amsmath}
\usepackage{amssymb}
\usepackage{amssymb}
\usepackage{wrapfig}
\usepackage{calligra}
\usepackage{subfig}
\usepackage{mathtools}
\usepackage{algorithm}
\usepackage{algorithmic}

\usepackage[T1]{fontenc}
\usepackage[utf8]{inputenc}
\usepackage{authblk}

\newcommand{\mcS}{\mathcal{S}}

\newcommand{\mcX}{\mathcal{X}}
\newcommand{\mcY}{\mathcal{Y}}
\newcommand{\mcL}{\mathcal{L}}
\newcommand{\mcZ}{\mathcal{Z}}
\newcommand{\mcQ}{\mathcal{Q}}
\newcommand{\mcC}{\mathcal{C}}
\newcommand{\mcN}{\mathcal{N}}
\newcommand{\mcM}{\mathcal{M}}

\newcommand{\mcD}{\mathcal{D}}
\newcommand{\mcJ}{\mathcal{J}}

\newcommand{\mbbR}{\mathbb{R}}
\newcommand{\mbbE}{\mathbb{E}}

\newcommand{\mbfx}{\mathbf{x}}
\newcommand{\mbfy}{\mathbf{y}}
\newcommand{\mbfz}{\mathbf{z}}

\newcommand{\mbfS}{\mathbf{S}}

\renewcommand{\and}{\mathrm{~and~}}

\newcommand{\mk}{m_k}
\newcommand{\mkk}{m_{k'}}

\newcommand{\mmk}{{m'}_k}
\newcommand{\mmkk}{{m'}_{k'}}
\newcommand{\mmmk}{{m''}_{k}}
\newcommand{\mmmmk}{{m'''}_{k}}
\newcommand{\mK}{m_K}

\newcommand{\balpha}{\boldsymbol\alpha}
\newcommand{\bmu}{\boldsymbol\mu}

\newcommand{\bSigma}{\boldsymbol\Sigma}

\newcommand{\bTheta}{\boldsymbol\Theta}
\newcommand{\bvTheta}{\boldsymbol\vartheta }

\newcommand{\argmin}{\arg\min}
\newcommand{\argmax}{\arg\max}

\begin{document}
\title{Clustering With Pairwise Relationships: \\A Generative Approach}

\author[1,2]{Yen-Yun Yu}
\author[2]{Shireen Y. Elhabian}
\author[1,2]{Ross T. Whitaker}
\affil[1]{School of Computing, University of Utah, Salt Lake City, UT 84108, USA}
\affil[2]{Scientific Computing and Imaging Institute, University of Utah, Salt Lake City, UT~84108, USA}

\date{}

\maketitle

\begin{abstract}
Semi-supervised learning (SSL) has become important 
in current data analysis applications, where the
amount of unlabeled data is growing exponentially 
and user input remains limited by logistics and expense.  
Constrained clustering, as a subclass of SSL, makes
use of user input in the form of \textit{relationships} between data points 
(e.g., pairs of data points belonging to the same class or different classes) 
and can remarkably improve the performance of unsupervised clustering 
in order to 
reflect user-defined knowledge of the relationships between particular data points.
Existing algorithms incorporate such user input, heuristically, as either 
hard constraints or soft penalties, which are separate from 
any generative or statistical aspect of the clustering model; 
this results in formulations that are suboptimal and not sufficiently general. 
In this paper, we propose a principled, {\em generative approach} to 
probabilistically model, without ad hoc penalties, the joint distribution given by user-defined pairwise {\em relations}. 
The proposed model accounts for general underlying distributions without assuming a specific form and relies on expectation-maximization for model fitting. 
For distributions in a standard form, the proposed approach results in a closed-form solution for updated parameters.
Results demonstrate the proposed model reflects user preferences with {\em fewer} user-defined relations compared to the state-of-the-art.

\end{abstract}

\section{Introduction}

Semi-supervised learning (SSL) has become a topic of significant
recent interest in the context of applied machine learning, 
where per-class distributions are difficult to
automatically separate due to limited sampling and/or limitations
of the underlying mathematical model.  
Several applications, including content-based retrieval~\cite{yang2012multimedia},
email classification~\cite{kyriakopoulou2013impact}, gene function prediction~\cite{nguyen2012detecting}, and natural language processing~\cite{sirts2013minimally,le2014semi}, 
benefit from the availability of user-defined/application-specific knowledge in the presence of 
large amounts of complex unlabeled data, where labeled
observations are often limited and expensive to acquire. 
In general, SSL algorithms fall into two broad categories: \textit{classification} and
\textit{clustering}. Semi-supervised classification is considered to
improve supervised classification when 
small amounts of labeled data with 
large amounts of unlabeled data are available
~\cite{zhu2006semi,chapelle2006semi}. For example, in a
semi-supervised email classification, one may wish to  classify constantly increasing email messages into spam/nonspam with the knowledge of a limited amount of user-/human-based classified messages~\cite{kyriakopoulou2013impact}. 
On the other hand, semi-supervised clustering (SSC), also known as {\it constrained clustering}~\cite{basu2008constrained},
aims to provide better performance for {\it unsupervised clustering} 
when user-based information about the \textit{relationships} within a small subset of the observations becomes available. Such relations would involve data points belonging to the same or different classes. 
For example, a language-specific grammar is necessary 
in cognitive science when individuals are attempting to learn a foreign language efficiently. 
Such a grammar provides rules for prepositions 
that can be considered as user-defined knowledge 
for improving the ability to learn a new language.

\begin{figure}
\center
\scalebox{.8}
{
    	\twoAcrossLabels {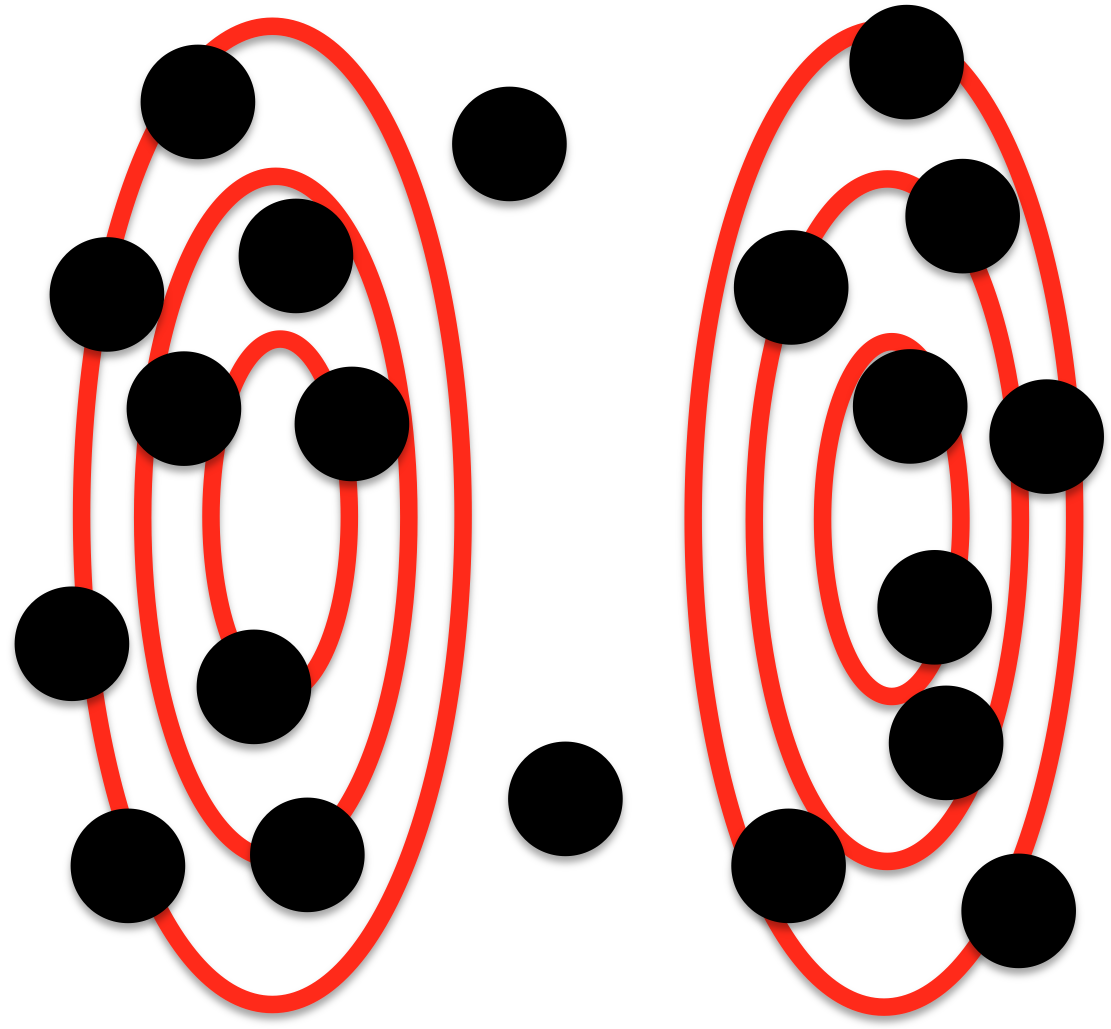}{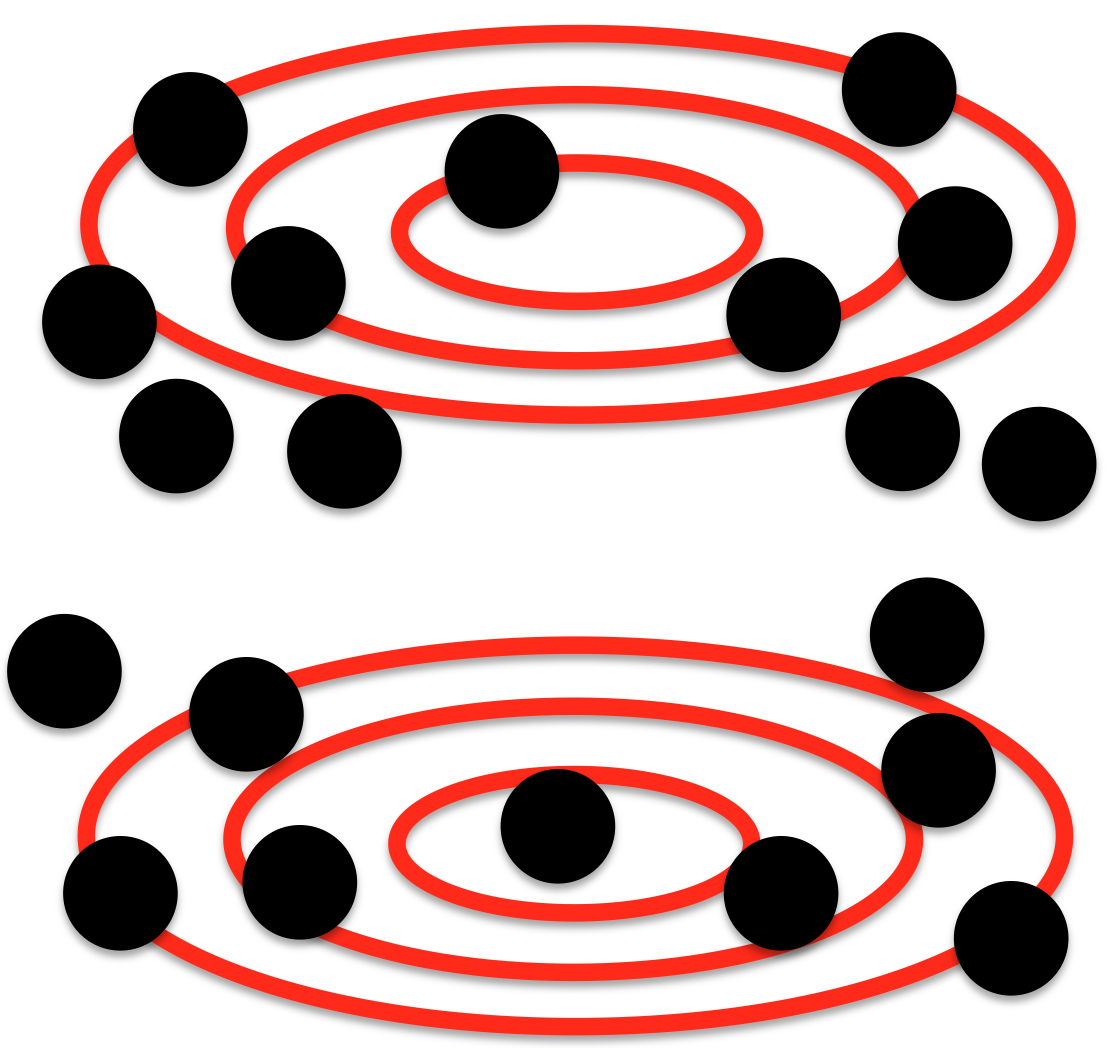}
	{(a) Mathematically Ideal Model} {(b) Application-Specific Model}
}
	\vspace{-5pt}
  	\caption
	[Generative model clustering example]
	{ {\bf  Generative model clustering example}: 
	 Because of finite sampling and modeling limitations, 
	 a distribution of points may give rise to optimal solutions that, depending on the model and the data, 
	 (a) are not well suited to the application and/or
	 (b) are not consistent with the underlying generative model, 
	 which may require domain knowledge from a user.
	}
 	\label{fig:wrongModel}
	\vspace{-5pt}
\end{figure}

To highlight the role of user-defined relationships for learning an application-specific data distribution, we consider the example in Figure
\ref{fig:wrongModel}(a), which shows a maximum likelihood model
estimate of a Gaussian mixture 
that is well supported by the data.  However, an application
may benefit from another good (but not optimal w.r.t. likelihood)
solution as in Figure~\ref{fig:wrongModel}(b), which is
inconsistent with the data, 
but is optimal without some information in addition to the raw data points.
Using a limited amount of
\textit{labeled} data and a large amount of unlabeled data could be
difficult to guide the learning algorithm in the application-specific
direction~\cite{zhu2006semi,cozman2003semi,loog2014semi,yang2011effect},
because performance of a generative model depends on the ratio 
of the labeled data to unlabeled data. 
In contrast, previous works have shown that SSC achieves the estimate
in Figure~\ref{fig:wrongModel}(b), given the observed data and a small
number of user-defined \textit{relationships} that would
\textit{guide} the parameter estimation process toward a
model~\cite{basu2008constrained} that is not only informed by the data,
but also by this small amount of user input.
This paper addresses the problem of incorporating such user-specific relations into a
clustering problem in an effective, general, and reliable manner.

Clustering data using a generative framework has some useful, important properties, 
including compact representations, parameter estimation for subsequent
statistical analysis, and the ability to induce classifications of unseen data~\cite{zhu2005harmonic}.
For the problem of estimating the parameters of generative models,
the expectation-maximization (EM)
algorithm~\cite{dempster1977maximum} is particularly effective.  
The EM formulation is guaranteed to give
maximum-likelihood (ML) estimates in the unimodal case and local
maxima in likelihood otherwise.  Therefore, EM formulations of
parameter estimation that properly account for user input in the
context of SSC are of interest and one of the contributions of this paper.

A flexible and efficient way to incorporate user input into SSC is in the form of
\textit{relations} between observed data points, in order to
define statistical relationships among observations (rather than
explicit labeling, as would be done in classification).  A typical example would be
for a user to examine a small subset of data and decide that some
pairs of points should be in different classes, referred to as a
\textit{cannot-link} relation, and that other pairs of data points should be in
the same class, i.e., \textit{must-link}. 
Using these basic primitives, one may build up more complex
relationships among sets of points.
The concept of pairwise links was first applied
to centroid-based clustering approaches, for instance, in the form of
\textit{constrained} K-means~\cite{wagstaff2001constrained}, where each
observation is assigned to the nearest cluster in a manner that avoids
violating constraints.

Although some progress has been made in developing mechanisms for
incorporating this type of user input into clustering algorithms, the need remains
for a systematic, general framework that generalizes with a limited amount of user knowledge.  
Most state-of-the-art techniques propose adding {\em hard constraints}~\cite{shental2004computing}, where data points that violate the constraints do not contribute
(i.e., all pairwise constraints must be satisfied), 
or 
{\em soft penalties}~\cite{lu2004semi}, 
which penalize the clustering results based on the number of violated constraints. 
Both hard constraints and soft penalties can lead to both a lack of generality and suboptimal solutions. 
For instance, in constrained K-means, introducing constraints by merely assigning a relatively
small number of points to appropriate centroids does not ensure that the models (centroids) adequately respond to this user input. 

In this paper, we propose a novel, generative approach for clustering
with pairwise relations that incorporates these relations into the
estimation process in a precise manner.  The parameters are estimated
by optimizing the data likelihood under the {\em assumption} that
individual data points are either independent samples (as in the
unsupervised case) or that they have a nontrivial joint distribution,
which is determined by user input.  The proposed model explicitly
incorporates the pairwise relationship as a property of the
generative model that guides the parameter estimation process to
reflect user preferences and estimates the global structure of the
underlying distribution.  Moreover, the proposed model is represented
as a probability distribution that can be virtually
any form. The results in this paper demonstrate
that the proposed optimal strategy pays off, and that it outperforms
the state-of-the art on real-world datasets with significantly less user input.

\section{Related Work}
\label{sec:relate}
Semi-supervised clustering methods typically fall into one of two categories~\cite{basu2008constrained}: \textit{distance-based} methods and
\textit{constraint-based} methods. The
distance-based approaches combine conventional clustering algorithms
with distance metrics that are designed to satisfy the information
given by user input
\cite{xing2002distance,bar2005learning,weinberger2005distance,cohn2003semi}.
The metrics effectively embed the points into spaces where the
distances between the points with constraints are either larger or
smaller to reflect the user-specified relationships. 
On the other hand,
constraint-based algorithms incorporate the pairwise
constraints into the clustering objective function, to either enforce
the constraints or penalize their violation. 
For example,Wagstaff et al. proposed the constrained
K-means algorithm, which enforced user input as hard constraints in a
nonprobabilistic manner as the part of the algorithm that assigns
points to classes~\cite{wagstaff2001constrained}. 
Basu el al. proposed a
probabilistic framework based on a hidden Markov random field, with
ad hoc soft penalties, which integrated metric learning with the
constrained K-means approach, optimized by an EM-like algorithm~\cite{basu2004probabilistic}.
This work also can be applied to a kernel feature space as in \cite{kulis2009semi}. 
Allab and Benabdeslem adapted topological
clustering to pairwise constraints using a self-organizing map in a
deterministic manner~\cite{allab2011constraint}.

Semi-supervised clustering methods with generative, parametric clustering approaches have also been augmented to accommodate user input.  
Lu and Leen proposed a penalized clustering algorithm using Gaussian mixture models (GMM)
by incorporating the pairwise constraints as a prior distribution over
the latent variable directly, resulting in a computationally challenging
evaluation of the posterior~\cite{lu2004semi}. Such a penalization-based formulation
results in a model with no clear generative interpretation 
and a stochastic expectation step that requires Gibbs sampling.
Shental et al. proposed a GMM
with equivalence constraints that defines the data from either
the same or a different source. However, for the {\em cannot-link} case, they used the Markov
network to describe the dependence between a pair of latent variables
and sought the optimal parameter by gradient ascent~\cite{shental2004computing}. Their
results showed that the cannot-link relationship was unable to 
impact the final parameter estimation (i.e., such a relation was ineffective). 
Further, they imposed user input as \textit{hard} constraints where data points that violate the constraints did not contribute to the parameter estimation process.  
A similar approach, in~\cite{law2005model}, proposed to treat the constraint as an additional
random variable that increases the complexity of the optimization process. Further, their approach focused only on {\em must-link}. 
In this paper, we propose a novel solution to incorporating user-defined data relationships into clustering problems, so that 
cannot-link and must-link relations can be included in a unified framework in a way that they are 
computed efficiently using an EM algorithm with very modest computational demands. 
Moreover, the proposed formulation is general in that it can 1) accommodate any kind of relation that can be expressed as a joint probability and 2) incorporate, in principle, any probability distribution (generative model).  
For GMMs, however, this formulation results in a particularly attractive algorithm that entails a closed-form solution for the mean and covariance and a relatively inexpensive, iterative, constrained, nonlinear optimization for the mixing parameters. 

Recently, EM-like algorithms for SSL
(and clustering in particular) have received significant attention in natural
language processing \cite{conf/nips/GracaGT07,mann2010generalized}.
Graca et al. proposed an EM approach 
with a posterior constraint that incorporates the expected values of
specially designed auxiliary functions of the latent variables to influence the
posterior distribution to favor user input~\cite{conf/nips/GracaGT07}.   Because of the lack of
probabilistic interpretation, the expectation step is not influenced by
user input, and the results are not optimal. 

Unlike the generative approach, 
graph-based methods group the data points according to similarity
and do not necessarily assume an underlying distribution.
Graph-based, semi-supervised clustering methods have 
been demonstrated to be promising when user input is available
~\cite{yi2013semi,wang2010flexible,xiong2012spectral}.
However, graph-based methods are not ideal classifiers 
when a new data point is presented 
due to their transductive property, 
i.e., unable to learn the general rule from the specific training data
~\cite{gammerman1998learning,zhu2005harmonic}. 
In order to classify a new data point, 
other than rebuilding the graph with the new data point, 
one likely solution is to build a separate inductive model on top of the output of the graph-based method
(e.g., K-means or GMM);
user input would need to be incorporated into this new model.


The work in this paper is distinct from the aforementioned works in the following aspects:
\begin{itemize}
	\item 
	 We present a {\em fully} generative approach, 
	rather than a heuristic approach of imposing hard constraints or adding ad hoc penalties.
	\item 
	The proposed generative model reflects user preferences
	while maintaining a probabilistic interpretation, which allows it to
	be generalized to take advantage of {\em alternative} density models or
	optimization algorithms. 
	\item 
	The proposed model clearly deals with the must-link {\em and} cannot-link
	cases in a unified framework
	and demonstrates that solutions using 
	must-link and cannot-link together or independently
	are tractable and effective.
	\item 
	Instead of pairwise constraints,
	the statistical interpretation of pairwise relationships 
	allows the model estimation to converge to a distribution 
	that follows user preferences with {\em less} domain knowledge.
	\item 
	In the proposed algorithm, the parameter estimation is
	very similar to a standard EM in terms of ease of implementation and efficiency. 
\end{itemize}


\section{Clustering With Pairwise Relationships}
\label{sec:FormulationSSC}
The proposed model incorporates user input 
in the form of relations between pairs of points that are 
in the same
class (\textit{must-link}) or 
different classes (\textit{cannot-link}). 
The \textit{must-link} and \textit{cannot-link} relationships are a natural and practical choice 
since the user can guide the clustering without having 
a specific preconceived notion of classes. 
These pairwise relationships are typically not sufficiently dense or
complete to build a full discriminative model, and yet they may be
helpful in discovering the underlying structure of the unlabeled data.
For data points that have no user input, we assume that they are
independent, random samples. 
The pairwise relationships give rise to an associate generative model with a
joint distribution that reflects the nature of the user input.

The parameters are estimated as an ML
formulation through an EM algorithm that
discovers the global structure of the underlying distribution that
reflects the user-defined relations.  
Unlike previous works that
include user input in a specific model (e.g., a GMM) through either hard constraints
~\cite{shental2004computing} or soft penalties
~\cite{lu2004semi}, 
in this work we propose an ML estimation
based on a generative model, without ad hoc penalties.

\subsection{Generative Models: Unsupervised Scenario}
\label{subsec:MM}
In this section, we first introduce generative models for an unsupervised scenario.
Suppose the unconstrained generative model consists of $M$ classes. 
$\mcX = \{\mbfx_n \in \mbbR^d\}_{n=1}^N$ denotes the observed dataset 
without user input.
Dataset $\mcX$ is associated with 
\textit{latent} set $\mcZ = \{\mbfz_n\}_{n=1}^N$ where
$\mbfz_n = [z_n^1, ..., z_n^M]^T \in \{0,1\}^M$ with $z_n^m = 1$ if
and only if the corresponding data point $\mbfx_n$ was generated from the $m$th class, 
subject to $\sum_{m=1}^M z_n^m = 1$.
Therefore, we can obtain the soft label for a data point $\mbfx$ by estimating $p(z^m|\mbfx)$. 
The probability that a data point $\mbfx$ is generated from a generative model 
with parameters $\bvTheta$ is
\begin{align}
	\label{eq:GMU}
		p(\mbfx | \bvTheta) 
		&
		=
		\sum_\mbfz   
		p(\mbfx | \mbfz, \bvTheta) 
		p(\mbfz)
\end{align}
The likelihood of the observed data points
governed by the model parameters is  
\begin{align}
	&
	\mcL(\mcX, \mcZ, \bvTheta) 
	:=
	p(\mcX, \mcZ| \bvTheta)
	= 
	\prod_{m=1}^M
	\prod_{n \in [1,N]: z_n^m=1} 	
	p(\mbfx_n) 
	\label{eqn:GM:prodconstraint}
	\\
	&= 
	\prod_{m=1}^M
	\prod_{n=1}^N 
	p(\mbfx_n, z_n^m) 
	=
	\prod_{m=1}^M 
	\prod_{n=1}^N 
	\bigg
	[ 
	p(\mbfx_n | z_n^m, \bvTheta)
	p(z_n^m)
	\bigg
	]^{z_n^m}
	\label{eqn:GM:jointprob}
\end{align}
where the condition on the product term in equation~(\ref{eqn:GM:prodconstraint}) is
restricted to data points $\mbfx_n$ generated from the $m$th class.
The joint probability in equation~(\ref{eqn:GM:jointprob}) is expressed, using Bayes' rule, in
terms of the conditional probability 
$p(\mbfx_n | z_n^m,\bvTheta)$
and the $m$th class prior probability  $p(z_n^m)$. 
In the rest of the formulation, to simplify the representation,  
we use $p(\mbfx_n | z_n^m) = p(\mbfx_n | z_n^m,\bvTheta)$
%
\subsection{Generative Model With Pairwise Relationships}
\label{subsec:GMPR}
\begin{figure*}[t!]
	\includegraphics[scale=0.18]{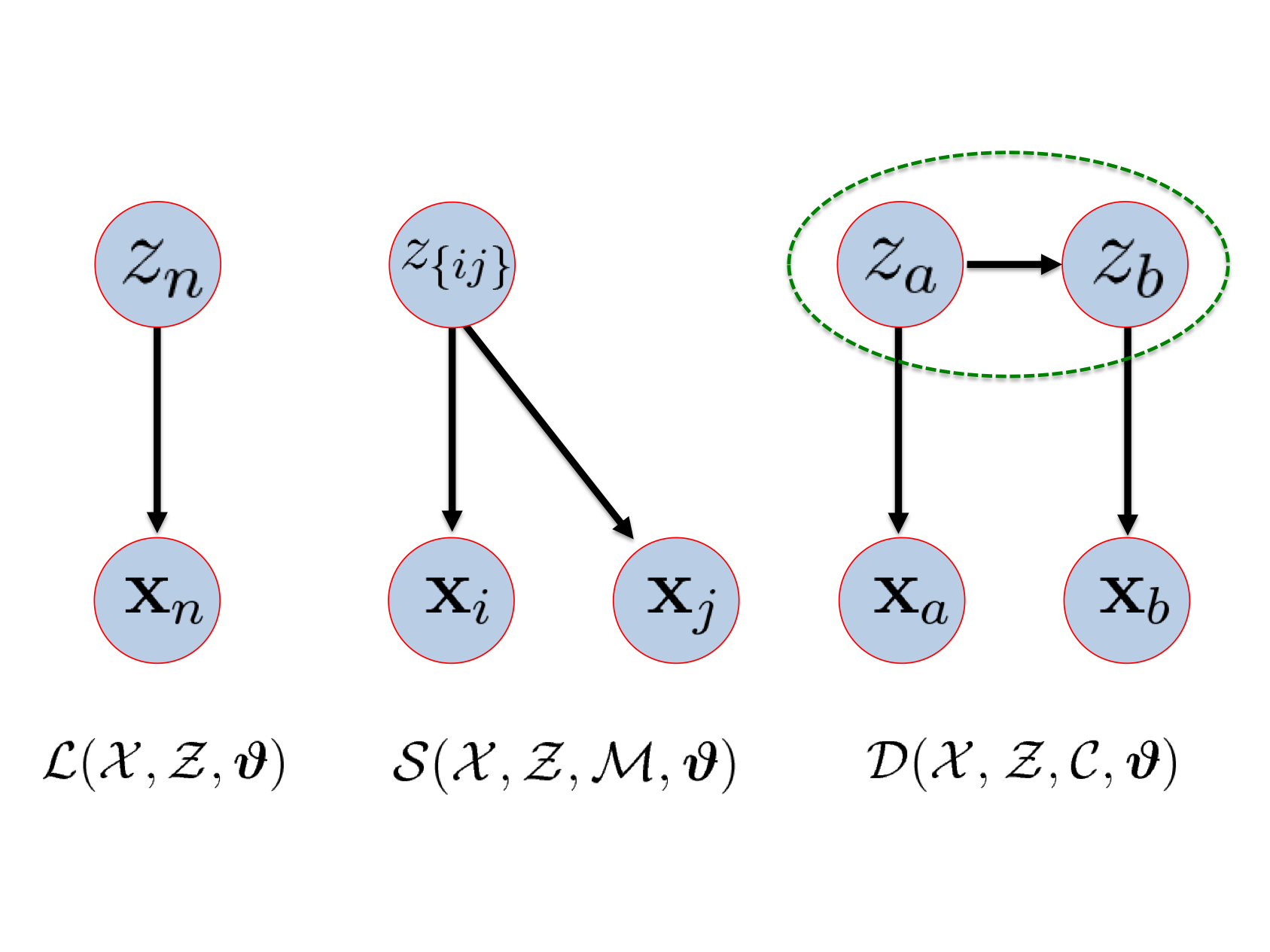}
	\centering
	\caption[Graphical representation of the generative model with pairwise relationships]
	{ 
		The graphical representation of the proposed generative model 
		with complete data-likelihood $p(\mcX, \mcZ | \mcM, \mcC,\bvTheta)$.
		The $\mcL(\cdot)$ is from the standard generative model with
                independent samples.  
		The $\mcS(\cdot)$ shows the must-link data points pair 
		$\mbfx_i$ and $\mbfx_j$ shares a single latent variable
                $z_{\{ij\}}$.
		The $\mcD(\cdot)$ shows the cannot-link data points pair
                $\mbfx_a$ and $\mbfx_b$, where
		the green dashed line indicates the joint probability
                of $z_a$ and $z_b$.
	}
 	\label{fig:graphModel}
\end{figure*}
%
%
The definition of a pairwise relation in the proposed generative model is similar to that in the unsupervised case, yet such relations are propagated to the latent variables level. In particular,  
$\mcM = \{(i,j)\}$ denotes a set of must-link relations where the pair $\mbfx_i$ and $\mbfx_j$ was generated from the same class; hence, the pair $(\mbfx_i,\mbfx_j)$ shares a single latent variable $\mbfz_{\{ij\}}$.
The same logic is applied to the cannot-link relations where $\mcC= \{(a,b)\}$ denotes a set of cannot-link relations encoding that
$\mbfx_a$ and $\mbfx_b$ were generated from distinct classes; therefore, $\mbfz_a \neq \mbfz_b$. 
Including $\mcM$ and $\mcC$, the data points are now expanded to be 
$\mcX := \{ \mbfx_1, \dots \mbfx_N, \mcM, \mcC \}$.
Thus, the \textit{modified} complete-data likelihood function $\mcJ(\cdot)$ that would reflect user input is (refer to Figure \ref{fig:graphModel} for the graphical representation)
\begin{align}
	\label{eq:LogJH}
	\mcJ(\mcX, \mcZ, \mcM, \mcC, \bvTheta) 
	&
	:= 
	p(\mcX, \mcZ |  \mcM, \mcC, \bvTheta) 
	\nonumber
	\\
	&
	=
	\mcL(\mcX, \mcZ, \bvTheta)~
	\mcS(\mcX, \mcZ, \mcM, \bvTheta)~
	\mcD(\mcX, \mcZ, \mcC, \bvTheta). 
\end{align}
$\mcS(\cdot)$ and $\mcD(\cdot)$ are the likelihood of pairwise data points.
The likelihood of the set of all pairs of must-link
data points $\mcS$ is, therefore, 
\begin{align}
	\mcS(\mcX,\mcZ, \mcM, \bvTheta) 
	&
	:=
	p(\mcX,\mcZ | \mcM, \bvTheta) 
	\nonumber
	\\
	&
	= 
	\prod_{m=1}^M
	\prod_{(i,j) \in \mcM} 
	p(\mbfx_i, \mbfx_j, z_{\{ij\}}^m) 
	\nonumber
	\\
	&
	=
	\prod_{m=1}^M 
	\prod_{(i,j) \in \mcM}
	\bigg
	[
	p(\mbfx_i| z_{\{ij\}}^m)
	p(\mbfx_j | z_{\{ij\}}^m)
	p(z_{\{ij\}}^m)
	\bigg
	]^{z_{\{ij\}}^m}
	\label{eq:Sterm}
\end{align}
%
The likelihood of the cannot-link data points explicitly 
reflects the fact that they are drawn from distinct classes.
Therefore, the joint probability of the labeling vectors 
$\mbfz_a$ and $\mbfz_b$ for all $(a,b) \in \mcC$ is as follows: 
\begin{eqnarray}
	p(z_a^m,z_b^m) &:=& p(z_a^m|z_b^m)p(z_b^m) 
	= 
	p(z_b^m|z_a^m)p(z_a^m)~~ 
	\label{eqn:jointCLbayes1}
	\\
	&=& \left\{
                \begin{array}{ll}
                  \cfrac
                  {p(z_a^m)^{z_a^m}p(z_b^m)^{z_b^m}}{1-\sum_{m'=1}^M p(z_a^{m'})^2} & z_a^m \neq z_b^m\\
                  0 & z_a^m = z_b^m \label{eqn:GM:propjoint}\\
                \end{array}
              \right. 
%
	\\
      &
      =
      &
      \cfrac
      {(1-z_a^m z_b^m) p(z_a^m)^{z_a^m}p(z_b^m)^{z_b^m}}
      {1-\sum_{m'=1}^M p(z_a^{m'})^2}
      \label{eq:joinCLbayes2}
\end{eqnarray} 
The proposed joint distribution reflects the
cannot-link constraints by assigning a zero joint probability of
$\mbfx_a$ and $\mbfx_b$ being generated from the same class,
and takes into account the effect of this relation on the
normalization term of the joint distribution $p(z_a^m,z_b^m)$.  
As such, the cannot-link relations contribute to the posterior distribution as follows:
\begin{align}
	\mcD(\mcX, \mcZ, \mcC, \bvTheta) 
	&
	:=
	p(\mcX, \mcZ | \mcC, \bvTheta) 
	\nonumber
	\\
	&
	= 
	\prod_{m=1}^M
	\prod_{(a,b) \in \mcC} 
	p(\mbfx_a, \mbfx_b,  z_a^m, z_b^m) 
	\nonumber
	\\
	&
	= 
	\prod_{m=1}^M
	\prod_{(a,b) \in \mcC}
	\Big
	[ 
	p(\mbfx_a| z_a^m)
	\Big
	]^{z_a^m}
	\Big
	[
	p( \mbfx_b| z_b^m) 
	\Big
	]^{z_b^m}
	p(z_a^m,z_b^m) 
	\label{eq:Dterm}
\end{align}
%
\subsection{Expectation Maximization With Pairwise Relationships}
\label{subsec:GMEM}

Given the joint distribution $p(\mcX, \mcZ | \mcM, \mcC, \bvTheta)$,
the objective is to maximize the log-likelihood function $\log \mcJ$ with
respect to the parameters $\bvTheta$ of the generative process in a
manner that would discover the global structure of the underlying
distribution and reflect user input. This objective
can be achieved using an EM algorithm.

\subsubsection{E-Step}
\label{subsubsec:Estep}
In the E-step, we estimate the posterior of the latent variables using the current parameter values
$\bvTheta^{\text{old}}$. 
\begin{align}
	\label{eq:Qfunction}
	\mcQ(\bvTheta,\bvTheta^{\text{old}}) 
	&
	= \mbbE_{\mcZ}[ \log ~ \mcJ ]
	\nonumber
	\\
	&
	=
	\sum_{\mcZ} p(\mcZ|\mcX,\mcM,\mcC,\bvTheta^{\text{old}}) 
	\log
	p(\mcX, \mcZ | \mcM, \mcC, \bvTheta)
\end{align}
\underline{$\mcL(\cdot)$-term:} Taking the expectation of $\log \mcL$ with respect to the
posterior distribution of $z_n^m$ and bearing in mind that the
latent variable $\mbfz$ is a binary variable, 
\begin{align}
	\label{eq:GM:Lterm}
	\mbbE_{z_n^m | \mbfx_n}[z_n^m] 
	= \frac
	{p(\mbfx_n | z_n^m)p(z_n^m)}
	{\sum_{m'=1}^M  p(\mbfx_n|z_n^{m'})p(z_n^{m'})}
\end{align}
\underline{$\mcS(\cdot)$-term:} Taking the expectation of $\log \mcS$ with respect to
the must-link posterior distribution of 
$z^m_{\{ij\}}$ results in
\begin{align}
	\label{eq:GM:Sterm}
	\mbbE_{z^m_{\{ij\}} | \mbfx_i, \mbfx_j}
	&
	[z^m_{\{ij\}}] 
	=
	\frac
	{
	p(\mbfx_i|z^m_{\{ij\}}) 
	p(\mbfx_j|z^m_{\{ij\}}) 
	p(z^m_{\{ij\}})
	}
	{
	\sum_{m'=1}^M 
	p(\mbfx_i|z^{m'}_{\{ij\}}) 
	p(\mbfx_j|z^{m'}_{\{ij\}})
	p(z^{m'}_{\{ij\}})
	}
\end{align}
\underline{$\mcD(\cdot)$-term:} Because the proposed model does not allow 
$\mbfx_a$ and $\mbfx_b$ to be from the same class, 
the expectation of equation~(\ref{eq:joinCLbayes2}) in the $\log \mcD$
that both will have the same class assignment vanishes,
which can be shown using Jensen's inequality as follows: 
\begin{align}
	\label{eq:GM:leq}
	\mbbE_{z^m_a, z^m_b | \mbfx_a, \mbfx_b}
	[\log(1-z_a^m z_b^m)] 
	&
	\leq
	\log 
	\big(
	1-\mbbE_{z^m_a, z^m_b | \mbfx_a, \mbfx_b}
	[z_a^m z_b^m]
	\big) 
	= \log(1-0) = 0
\end{align}
Hence, we can set $\log(1-z_a^m z_b^m) = 0$ 
in equation~(\ref{eq:joinCLbayes2}). 
The expectation of the $\log \mcD$ term with respect to 
$z^m_a$ is
\begin{align}
	\label{eq:GM:Dterm:z}
	\mbbE_{z^m_a | \mbfx_a, \mbfx_b}
	[z_a^m] 
	&
	= 
	p(z_a^m| \mbfx_a, \mbfx_b) 
	=
	\sum_{m'=1}^M
	p(z_a^m, z_b^{m'} | \mbfx_a, \mbfx_b)
	\nonumber
	\\
	&
	= 
	\frac
	{ 
	\sum_{m'=1}^{M}
	p(\mbfx_a|z_a^m)
	p(\mbfx_b|z_b^{m'})
	p(z_a^m,z_b^{m'})
	}
	{
	\sum_{m''=1}^{M}
	\sum_{m'''=1}^{M}
	p(\mbfx_a|z_a^{\mk''})
	p(\mbfx_b|z_b^{\mk'''})
	p(z_a^{m''},z_b^{m'''})
	}
	.
\end{align}
In a like manner, 
we can write down the expectation of $z_b^m$.
\subsubsection{M-Step}
\label{subsubsec:Mstep}
In the M-step, therefore, we update the $\bvTheta^{\text{new}}$ by maximizing equation~(\ref{eq:Qfunction}) and fixing the posterior distribution that we estimated in the E-step.
 \begin{align}
	\label{eq:argmaxQ}
	\bvTheta^{\text{new}} = \argmax_{\bvTheta} ~~ \mcQ(\bvTheta,\bvTheta^{\text{old}})
\end{align}
Different density models result in different update mechanisms for the respective model parameters. 
In the next subsection, we elaborate on an example of the proposed model to illustrate the idea of the M-step for the case of Gaussian mixture models.

\subsection{Gaussian Mixture Model With Pairwise Relationships}
\label{subsubsec:GMM}
Consider employing a single distribution 
(e.g., a Gaussian distribuion) 
for each class probability $p(\mbfx|z^m)$.
The proposed model, therefore, becomes the Gaussian mixture model (GMM)
with pairwise relationships. 
The parameter of the GMM is $\bvTheta = \{\alpha_m, \bmu_m, \bSigma_m \}_{m=1}^M$,
such that $\alpha_m \in [0,1]$ is the mixing parameter for the {\em class} proportion subject to 
$\sum_{m=1}^M \alpha_m =1$ and 
$p(z^m) = \alpha_m$. 
$\bmu_m \in \mbbR^d$ is the mean parameter, and 
$\bSigma_m \in \mbbR^{d \times d}$ is the covariance associated with the $m$th class.
By taking the derivative of equation~(\ref{eq:Qfunction}) with respect to $\bmu_m$ and $\bSigma_m$,
we can get
\begin{align}
	\label{eqn:mu}
	\bmu_m
	&
	= 
	\bigg ( 
	\sum_{n=1}^N \ell_n^m \mbfx_n 
	+ 
	\sum_{(i,j) \in \mcM} s_{ij}^m
	\big[
	\mbfx_i + \mbfx_j
	\big] 
	+ 
	\sum_{(a,b) \in \mcC}
	\big[
	d_{a}^m\mbfx_a + d_{b}^m\mbfx_b
	\big]
	\bigg )
	\bigg / Z
	\\
	\label{eqn:cov}
	\bSigma_m 
	&
	= 
	\bigg (
	\sum_{n=1}^N \ell_n^m \mbfS_n^m 
	+ 
	\sum_{(i,j) \in \mcM} s_{ij}^m
	\big[
	\mbfS_i^m + 
	\mbfS_j^m
	\big] 
	+ 
	\sum_{(a,b) \in \mcC}
	\big [
	d_{a}^m\mbfS_a^m + d_{b}^m \mbfS_b^m
	\big ]
	\bigg )
	\bigg / Z
	\\
	\label{eqn:normalization}
	Z 
	&
	= 
	\sum_{n=1}^N \ell_n^m + 2\sum_{(i,j) \in \mcM} s_{ij}^m + \sum_{(a,b) \in \mcC}
	\big[
	d_{a}^m + d_{b}^m
	\big]
\end{align}
where 
$\ell_n^m = p(z_n^m|\mbfx_n)$, $s_{ij}^m = p(z_{\{ij\}}^m|\mbfx_i, \mbfx_j)$, $d_{a}^m = p(z_a^m|\mbfx_a, \mbfx_b)$ and the sample covariance $\mbfS_n^m = (\mbfx_n - \bmu_m)(\mbfx_n - \bmu_m)^T$.

Estimating the mixing parameters $\alpha_m$, on the other hand, entails the
following constrained nonlinear optimization, which can be solved
using sequential quadratic programming with Newton-Raphson steps
\cite{fletcher2013practical,abramowitz1964handbook}. 
Let $\balpha \in \mbbR^M$ denote the vector
of mixing parameters. Given the current estimate of the mean  vectors
and covariance matrices, the new estimate of the mixing parameters can
be solved for using the optimization problem defined in (\ref{eqn:mixing_opt}),
\begin{align}
	\label{eqn:mixing_opt}
	\balpha^* &= \argmin_{\balpha} -\mcQ(\bvTheta,\bvTheta^{\text{old}}) 
	\nonumber
	\\
	\operatorname{s.t.}& 
	~~~~~
	\mathbf{1}^T\balpha - 1 = 0 ~~\operatorname{and}~~ \alpha_m \geq 0 ~~\forall m \in [1,M]
\end{align}
where the initialization can be obtained using the closed-form
solution obtained from discarding the nonlinear part, which ignores
the normalization term $\log(1-\sum_{m'=1}^M \alpha_{m'}^2)$. The energy
function is convex, and we have found that this
iterative algorithm typically converges in three to five
iterations and does not represent a significant computational burden.


\subsubsection{Multiple Mixture Clusters Per Class}
\label{subsec:MM}
In order to group the data that lies on the subspace 
(e.g., manifold structure) more explicitly,
multiclusters to model per class have been widely used in unsupervised clustering
by representing the density model in a hierarchical structure
~\cite{coviello2012variational,williams1999mcmc,vasconcelos1998learning,goldberger2004hierarchical,meila2001learning,jordan1994hierarchical,titsias2002mixture}.
Because of its natural representation of data, 
the hierarchical structure can be built using either a top-down or bottom-up approach,
in which the first approach tries to decompose one cluster into several small clusters,
whereas the second starts with grouping several clusters into one cluster.
The multicluster per class strategy also has been proposed 
when both labeled data and unlabeled data are available
~\cite{nigam2000text,liu2010gaussian,shen2012refining,he2011laplacian,xing2013multi,demiriz1999semi,dara2002clustering,goldberg2009multi}.
However, previous works indicated
that the labeled data is unable to impact the final parameter estimation if the initial model assumption is incorrect~\cite{cozman2003semi,loog2014semi,yang2011effect,singh2009unlabeled}.
Moreover, it is not clear how to employ the previous works in regard to pairwise links instead of labeled data.

In this section, we propose to use the generative 
mixture of Gaussian distributions 
for each class probability $p(\mbfx|z^m)$.
In this form, we use multiclusters to model one class that 
overcomes data on a manifold structure.
Therefore, in addition to the latent variable set $\mcZ$,
$\mcX$ is also associated with the 
\textit{latent} variable set $\mcY = \{\mbfy_n\}_{n=1}^N$ where
$\mbfy_n = [y_n^1, ..., y_n^{\mK}]^T \in \{0,1\}^{\mK}$ with $y_n^{\mk} = 1$ if
and only if the corresponding data point $\mbfx_n$ was generated from 
the $k$th cluster in the $m$th class, 
subject to $\sum_{\mk=1}^{\mK} y_n^{\mk} = 1$; 
$\mK$ is the number of clusters in the $m$th class.
The parameter of the generative mixture model is 
$\bvTheta = \{\alpha_m, \bTheta_m\}_{m=1}^M$ and 
$\alpha_m$ is the mixing parameter for the {\em class} proportion 
and is the same as $\alpha_m$ in section~\ref{subsubsec:GMM}.
The parameter of the $m$th class is 
$\bTheta_m = \{\pi_{\mk}, \Theta_{\mk}\}_{\mk=1}^{\mK}$ where
$\Theta_{\mk} = \{\bmu_{\mk}, \bSigma_{\mk}\}$, 
such that $\pi_{\mk} \in [0,1]$ is the
mixing parameter for the {\em cluster} proportion subject to 
$\sum_{\mk=1}^{\mk} \pi_{\mk}=1$, 
$\bmu_{\mk} \in \mbbR^d$ is the mean parameter, and 
$\bSigma_{\mk} \in \mbbR^{d \times d}$ is the covariance associated with the
$k$th cluster in the $m$th class.
The probability that an unsupervised data point $\mbfx$ is generated from a generative mixture model 
given parameters $\bvTheta$ is
\begin{align}
		&
		\mcL(\mcX, \mcY, \mcZ, \bvTheta) 
		=
		\prod_{m=1}^M 
		\prod_{\mk=1}^{\mK} 
		\prod_{n=1}^N 
		\bigg
		[	 
		\Big
		[
		p(\mbfx_n | y_n^{\mk})
		p(y_n^{\mk}|z_n^m)
		\Big
		]^{y_n^{\mk}}
		p(z_n^m) 
		\bigg
		]^{z_n^m}
		\label{eqn:HGMM:jointprob}
\end{align}
where
\begin{align}
		p(z_n^m) 	
		=
          \alpha_m
          ;~
		p(y_n^{\mk} | z_n^m) 
		= 
		\pi_{\mk}
          ;~
           p
		(\mbfx_n| y_n^{\mk}) 
		= 
		\mcN (\mbfx_n|\bmu_{\mk}, \bSigma_{\mk})
		,
		\label{eq:HGMM:individual}
\end{align}
and $\mcN (\mbfx_n|\bmu_{\mk}, \bSigma_{\mk})$ is the Gaussian distribution.
The definition of equation~(\ref{eq:HGMM:individual})  
can be used to describe the 
$\mcS(\cdot)$ in equation~(\ref{eq:Sterm}) and the
$\mcD(\cdot)$ in equation~(\ref{eq:Dterm}). 
In the E-step, 
the posterior of latent variable $\mcZ$ can be estimated by 
marginalization of the $\mcY$ directly.
In the M-step,
we update the parameters by maximizing equation~(\ref{eq:Qfunction}),
which is similar to GMM case in section~\ref{subsubsec:GMM} 
(see the Appendix A for details).
Last, if $\mK=1$, we have $\mbfy_n = [y_n^1]$ and equation~(\ref{eqn:HGMM:jointprob}) becomes the GMM, 
i.e., one cluster/single Gaussian distribution per class.

\section{Experiment}
\label{sec:experiment}
In this section, we demonstrate the effectiveness of the proposed generative model on a synthetic dataset as well as on well-known datasets where the number of links can be significantly reduced compared to state-of-the-art.

\subsection{Experimental Settings}
\label{subsec:initialization}
To illustrate the method,
 we start with the case of $p(\mbfx|z^m)$: 
 a mixture of Gaussians ($\mK > 1$) and 
a single Gaussian distribution ($\mK = 1$). 
To initialize the model parameters, we first randomly select the mean vectors by K-means++~\cite{arthur2007k}, which is similar to the Gonzalez algorithm~\cite{gonzalez1985clustering} without being completely greedy.
Afterward, we assign every observed data point to its nearest initial mean where initial covariance matrices for each class are computed. 
We initially assume equally probable classes where the mixing parameters are set to $1/M$. 
When $\mK > 1$ (i.e., multiclusters per class), 
we initialize the parameters of the $k$th cluster in the $m$th class using
the aforementioned strategy, but only on the data points that have been assigned to 
the $m$th class after the above initialization.
To mimic user preferences and assess the performance of the proposed model as a function of the number of available relations, pairwise relations are created by randomly selecting a pair of observed data points and using the knowledge of the distributions. If the points are assigned to the same cluster based on their ground truth labeling, we move them to the {\em must-link} set, otherwise, to the {\em cannot-link} set. 
We perform 100 trials for all experiments. 
Each trial is constructed by the random initialization of the model parameters 
and random pairwise relations.

We compare the proposed model, \textit{generative model with pairwise relation}
({\bf GM-PR}), to the unconstrained {\bf GMM}, unconstrained {\bf spectral clustering (SC)}, and four other state-of-the-art algorithms:
1) {\bf GMM-EC}: GMM with the equivalence constraint~\cite{shental2004computing},
2) {\bf EM-PC}: EM with the posterior constraint~\cite{conf/nips/GracaGT07}; it is worth mentioning that {\bf EM-PC} works only for {\em cannot-link},
3) {\bf SSKK}: Constrained kernel K-means~\cite{kulis2009semi}, and
4) {\bf CSC}~\footnote{https://github.com/gnaixgnaw/CSP}: 
Flexible constrained spectral clustering~\cite{wang2010flexible}. 
For SC, SSKK, and CSC, the similarity matrix is computed by the RBF kernel, whose parameter is set by the average squared distance between all pairs of data points. 

We use {\em purity}~\cite{manning2008introduction} for performance evaluation, which is a scalar value ranging from $0$ to $1$ where $1$ is the best. 
Purity can be computed as follows: 
each class $m$ is assigned to the most frequent ground truth label $g(m)$; then, purity is measured by counting the number of correctly assigned observed data points in every ground truth class and dividing the total number of observed data. The assignment is according to the highest probability of the posterior distribution. 

\subsection{Results: Single Gaussian Distribution ($\mK=1$)}
In this section, we demonstrate the performance of the proposed model 
using a single Gaussian distribution 
on standard binary and multiclass problems.

\label{subsec:G}
\subsubsection{Synthetic Data}
\label{subsubsec:G:toy}
\begin{figure*}[!t]
	\twoAcrossLabels{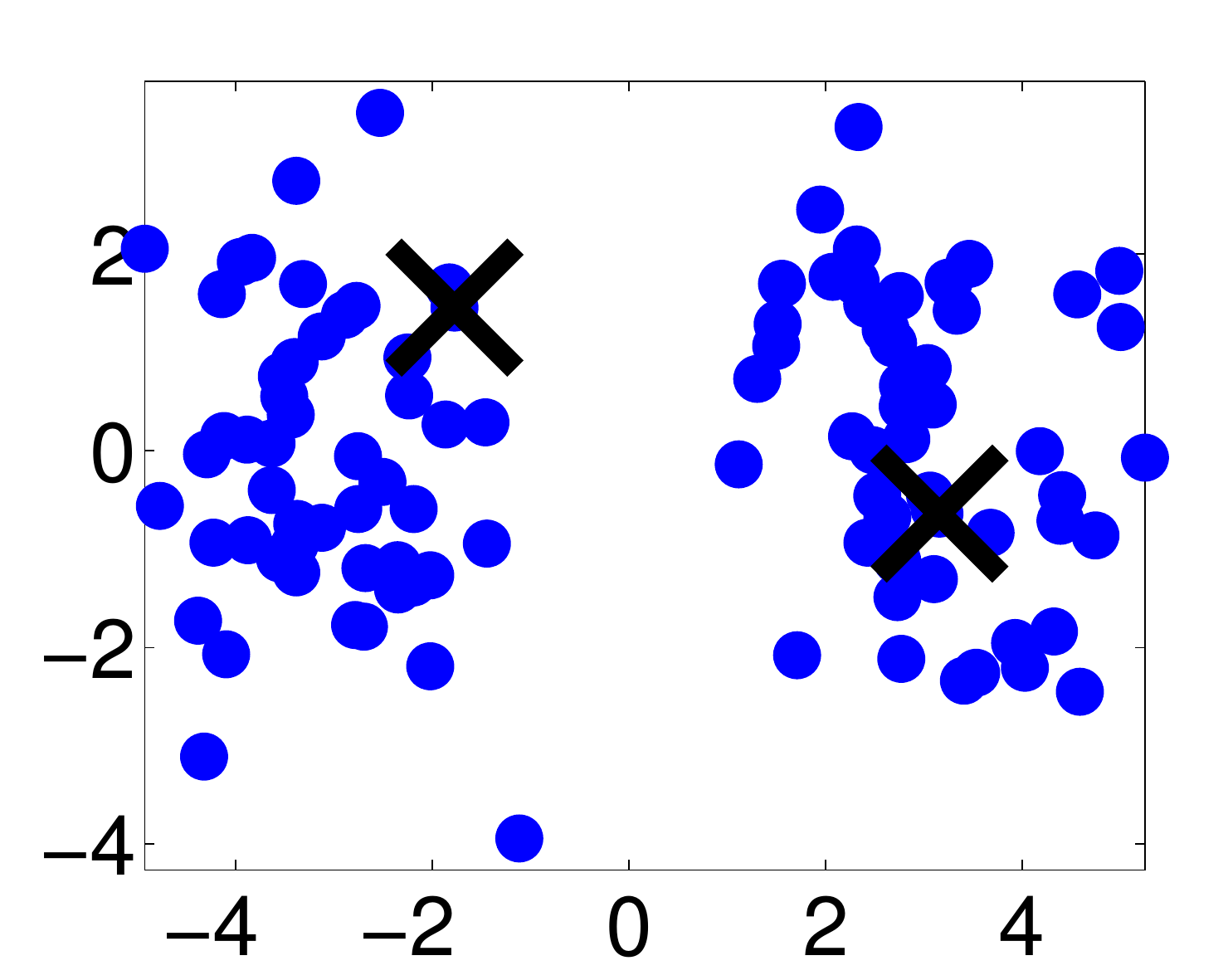}{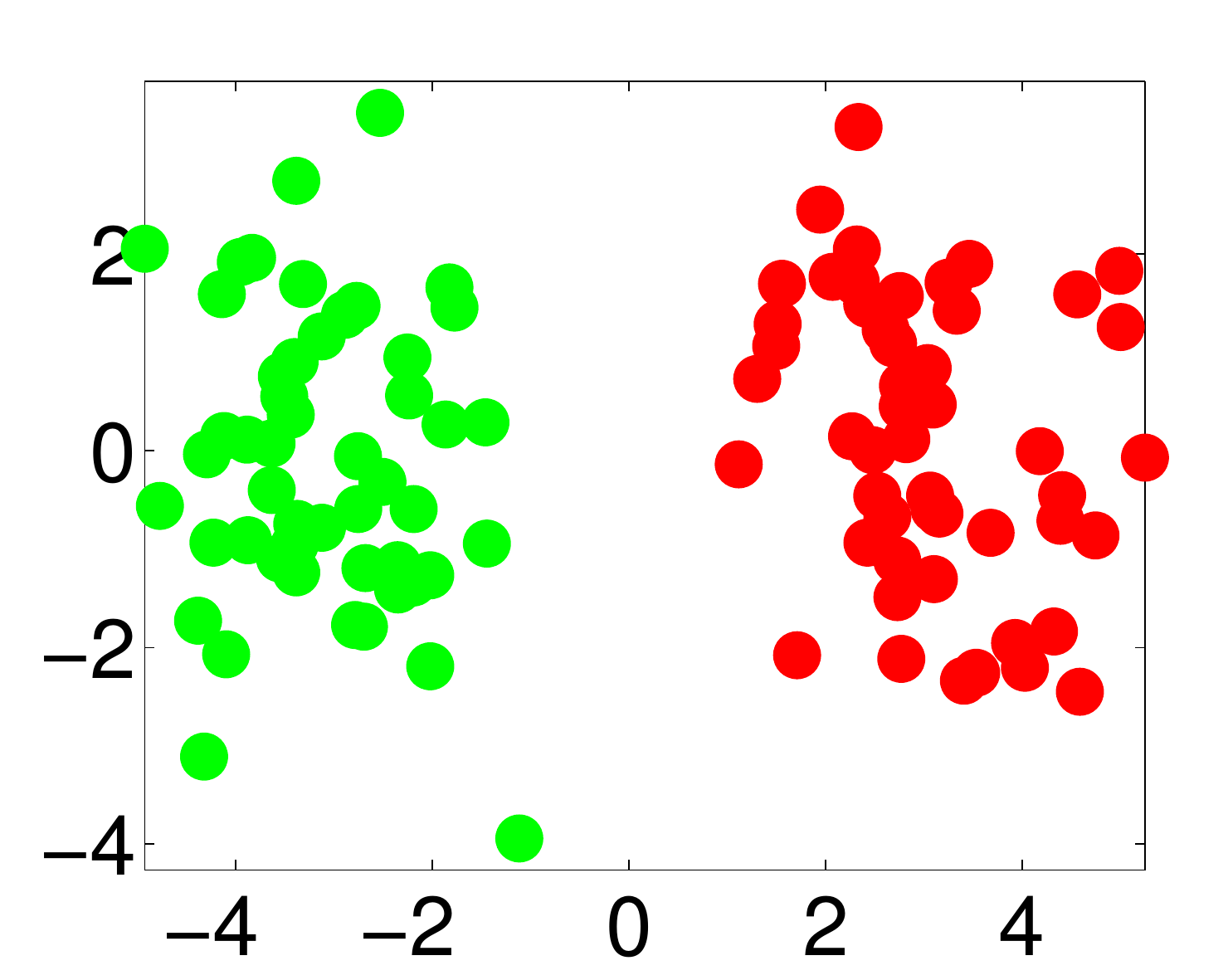}
	{(a)}{(b)}
	\threeAcrossLabels{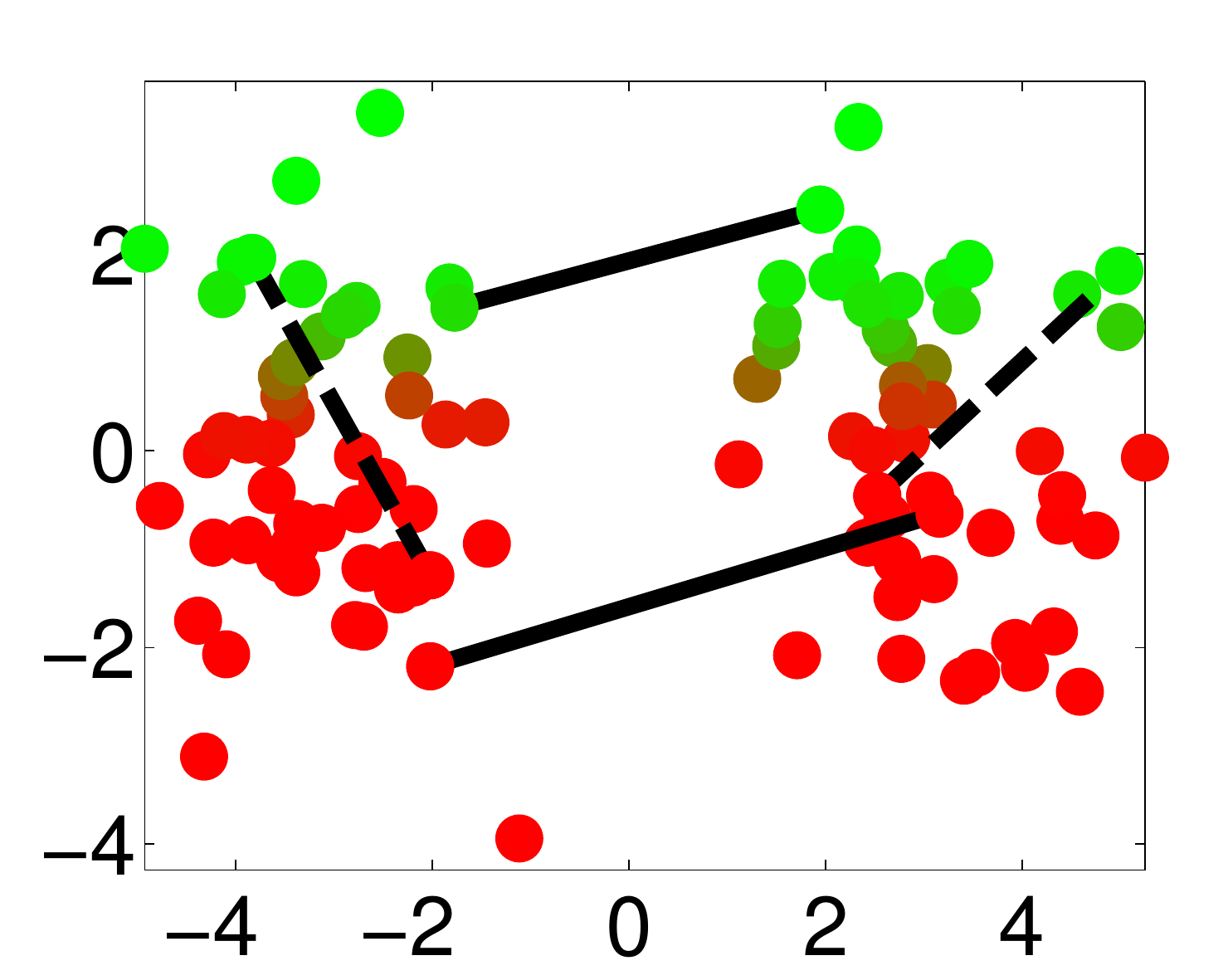}{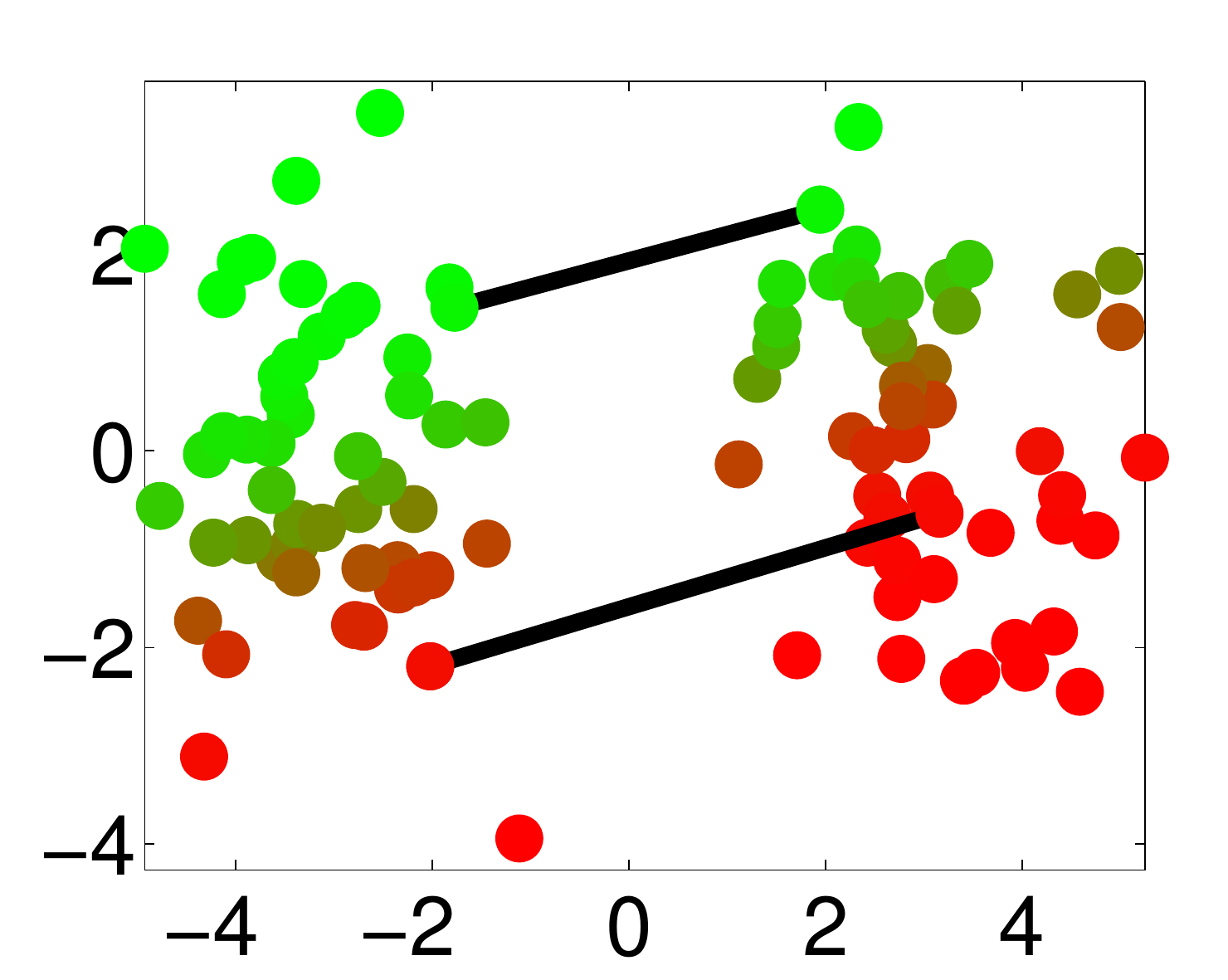}{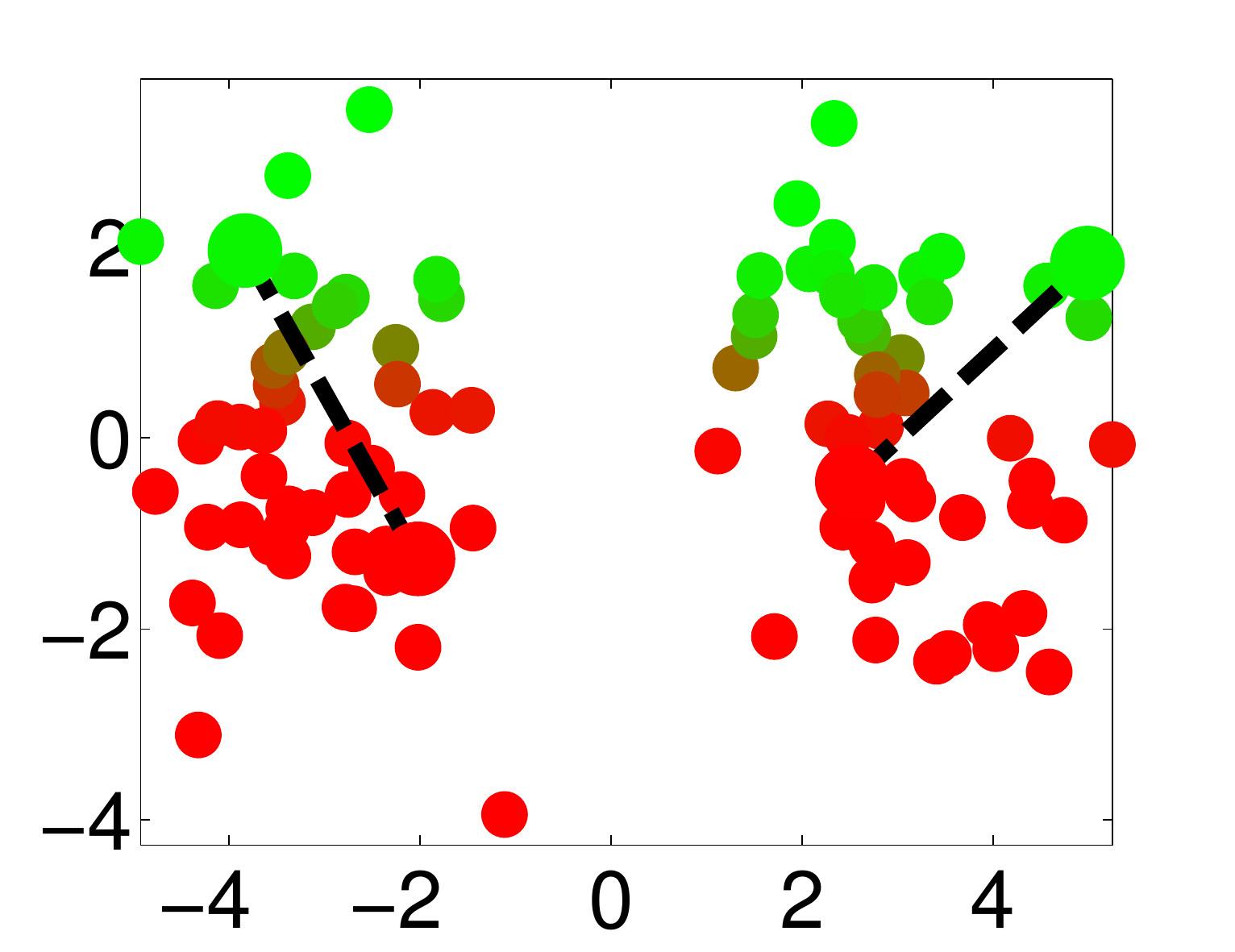}
	{(c)}{(d)}{(e)}	
     \vspace{-5pt}
	\caption
	[Result of application-specific model synthetic data]
	{ {\bf  Application-specific model synthetic data}: 
	(a) Original data with initial two means marked by x.
	Results are represented as follows:
	(b) {\bf GMM}, 
	(c) {\bf GM-PR} using two {\em must-links} (solid line)
	and two {\em cannot-links} (dashed line),
	(d) {\bf GM-PR} using only two must-links, and 
	(e) {\bf GM-PR} using only two cannot-links.
	The saturation of the red/green points represents the value of the soft label.
	}
 	\label{fig:toy}
	\vspace{-5pt}
\end{figure*}

We start off by evaluating the performance of {\bf GM-PR},
which uses a single Gaussian distribution for $p(\mbfx|z^m)$ on synthetic data. 
We generate a two-cluster toy example to mimic the example in Figure~{\ref{fig:wrongModel}}, which is motivated by \cite{zhu2006semi}. 
The correct decision boundary should be the horizontal line along the x-axis. 
Figure~\ref{fig:toy}(a) is the generated data with the initial means. 
Figure~\ref{fig:toy}(b) is the clustering result obtained from an unconstrained {\bf GMM}. 
Figure~\ref{fig:toy}(c) shows that the proposed \textbf{GM-PR} can learn the desired model 
with only two must-link relations and two cannot-link relations. 
Figure~\ref{fig:toy}(d) shows that the proposed \textbf{GM-PR} can learn the desired model 
with only two must-links. 
Figure~\ref{fig:toy}(e) shows that the proposed \textbf{GM-PR} can learn the desired model 
with only two cannot-link relations. 
This experiment illustrates the advantage of the proposed method, 
which can perform well with only either must-links or cannot-links.
This advantage makes the proposed model distinct from previous works 
~\cite{shental2004computing,law2005model}.

\subsubsection{UCI Repository and Handwritten Digits}
\label{subsubsec:G:RealData}
In this section, we report the performance of three real datasets:
1) the {\bf Haberman's survival}\footnote{https://archive.ics.uci.edu/ml/datasets.html}
  	 dataset contains 306 instances, 3 attributes, and 2 classes;
2) the {\bf MNIST}\footnote{http://yann.lecun.com/exdb/mnist/} 
  	database contains images of handwritten digits. We used the test dataset, 
  	which contains 10000 examples, 784 attributes, and 10 classes~\cite{lecun1998gradient}; and 
3) the {\bf Thyroid}\footnote{http://www.raetschlab.org/Members/raetsch/benchmark}
  	dataset contains 215 instances, 5 attributes, and 2 classes.

\begin{figure*}[!t]	
	\center
	\begin{minipage}[b]{0.1\linewidth}
		\includegraphics[width=2.1cm,height=1.4cm]{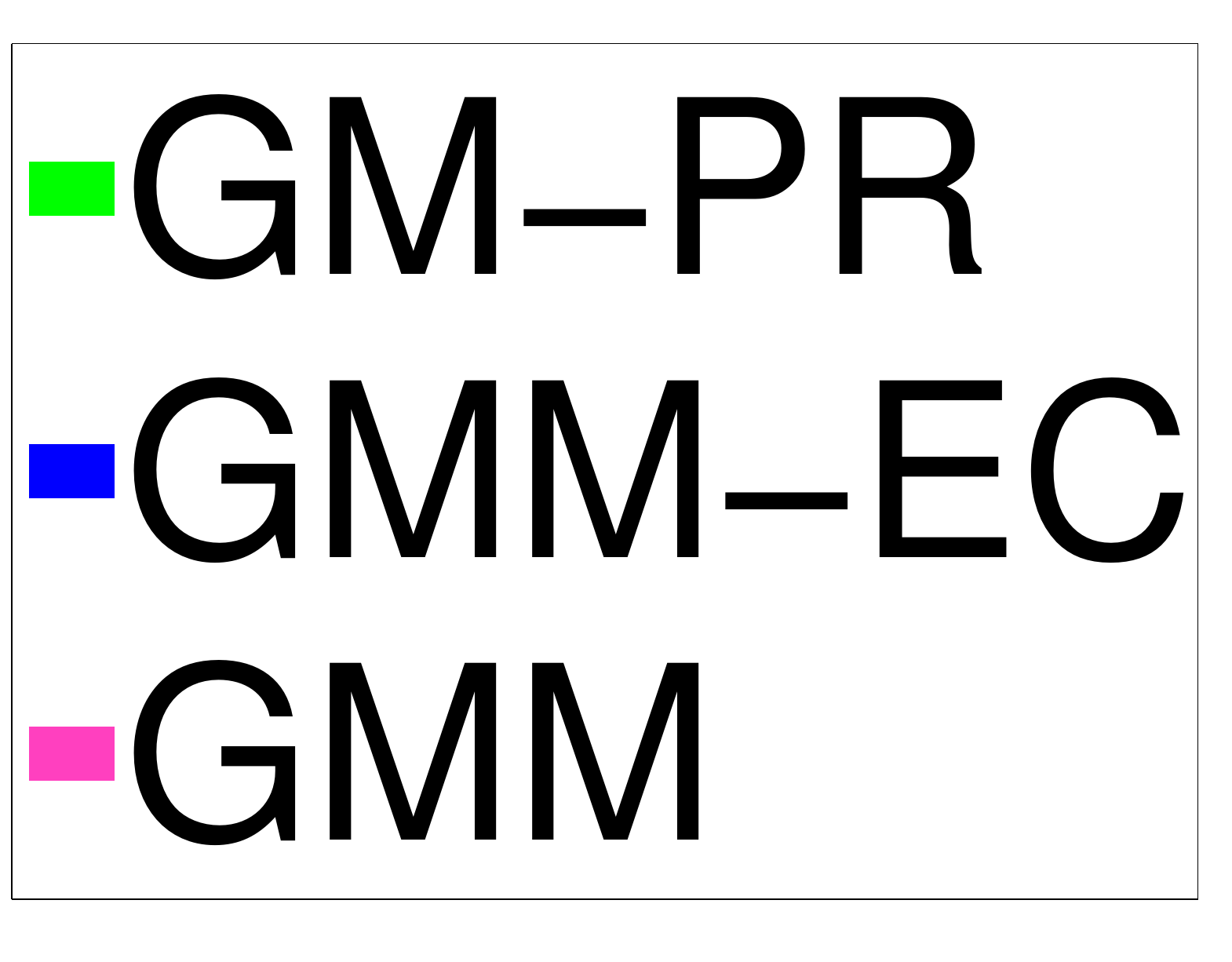}
	\end{minipage}%
	\begin{minipage}[t]{\linewidth}
		\twoAcrossLabels{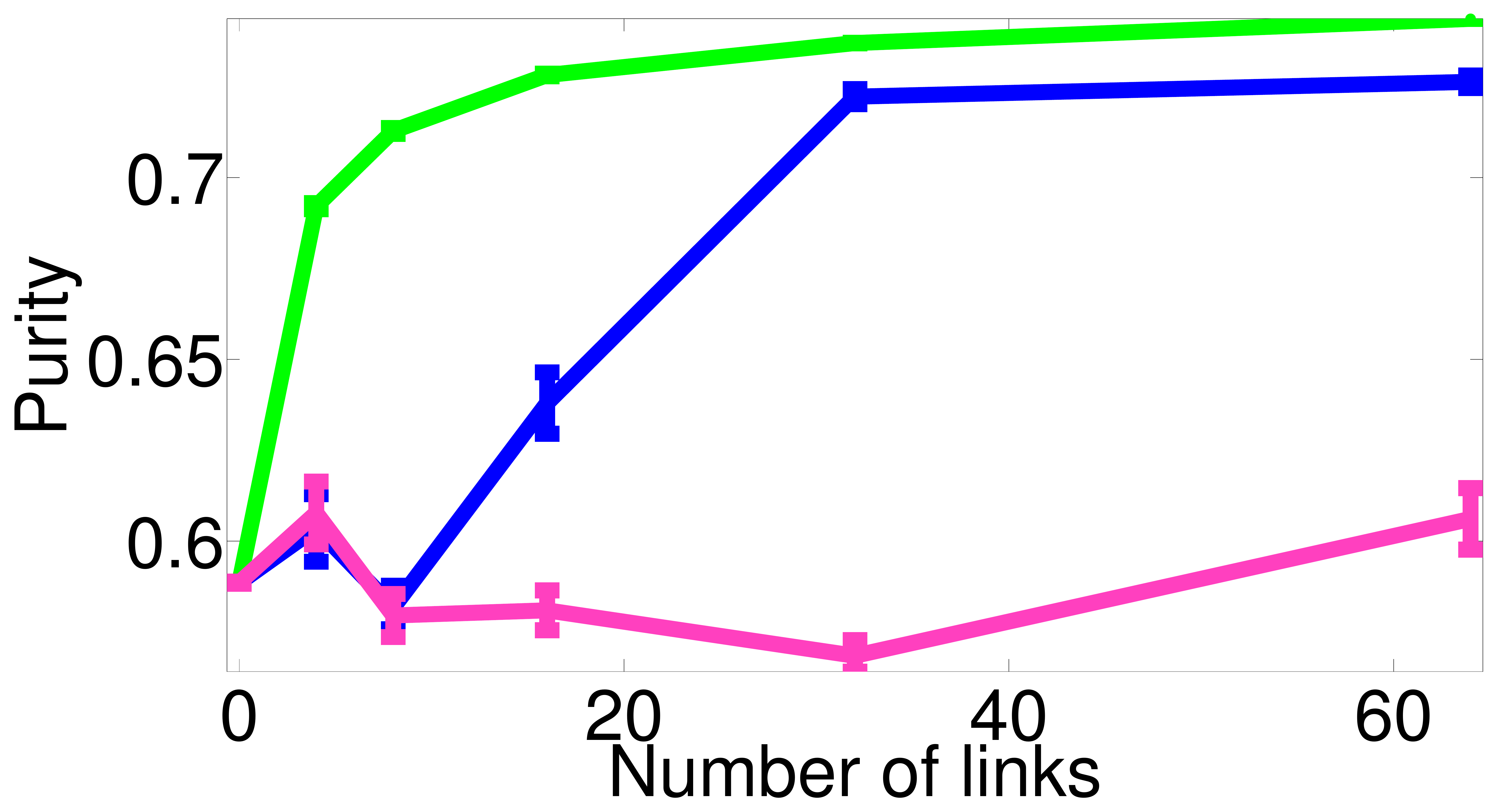}{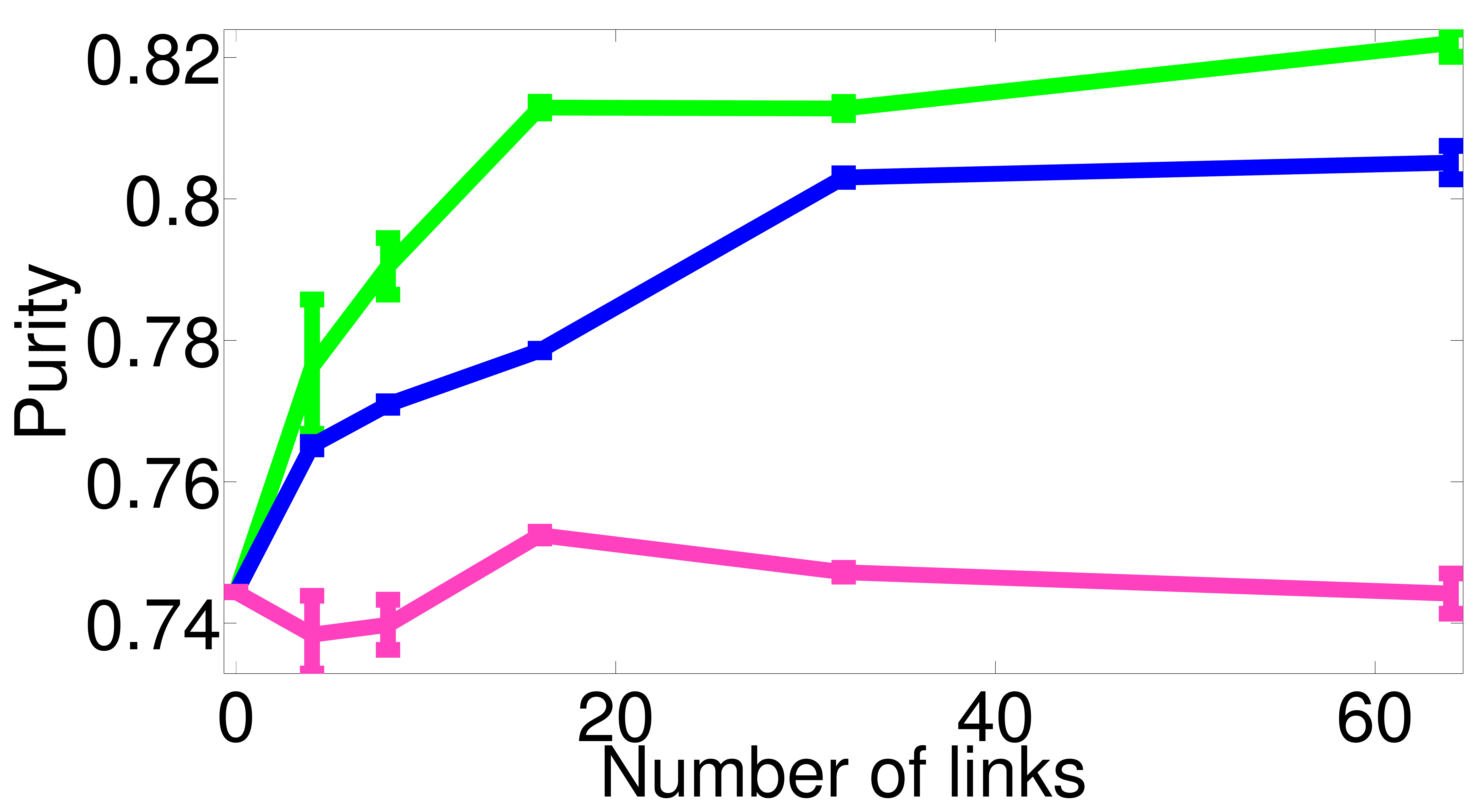}
		{(a) Harberman}{(b) Thyroid}
		\twoAcrossLabels{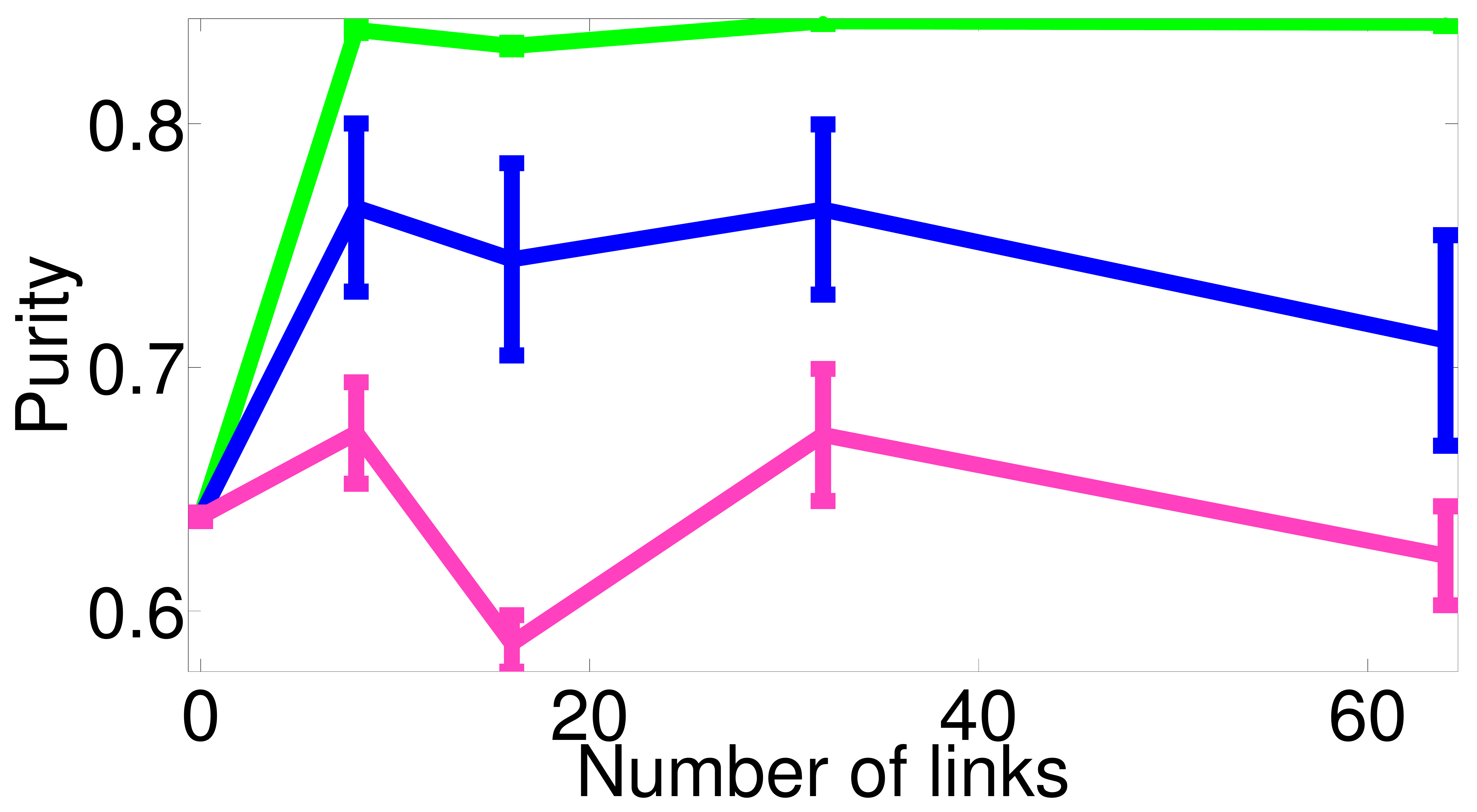}{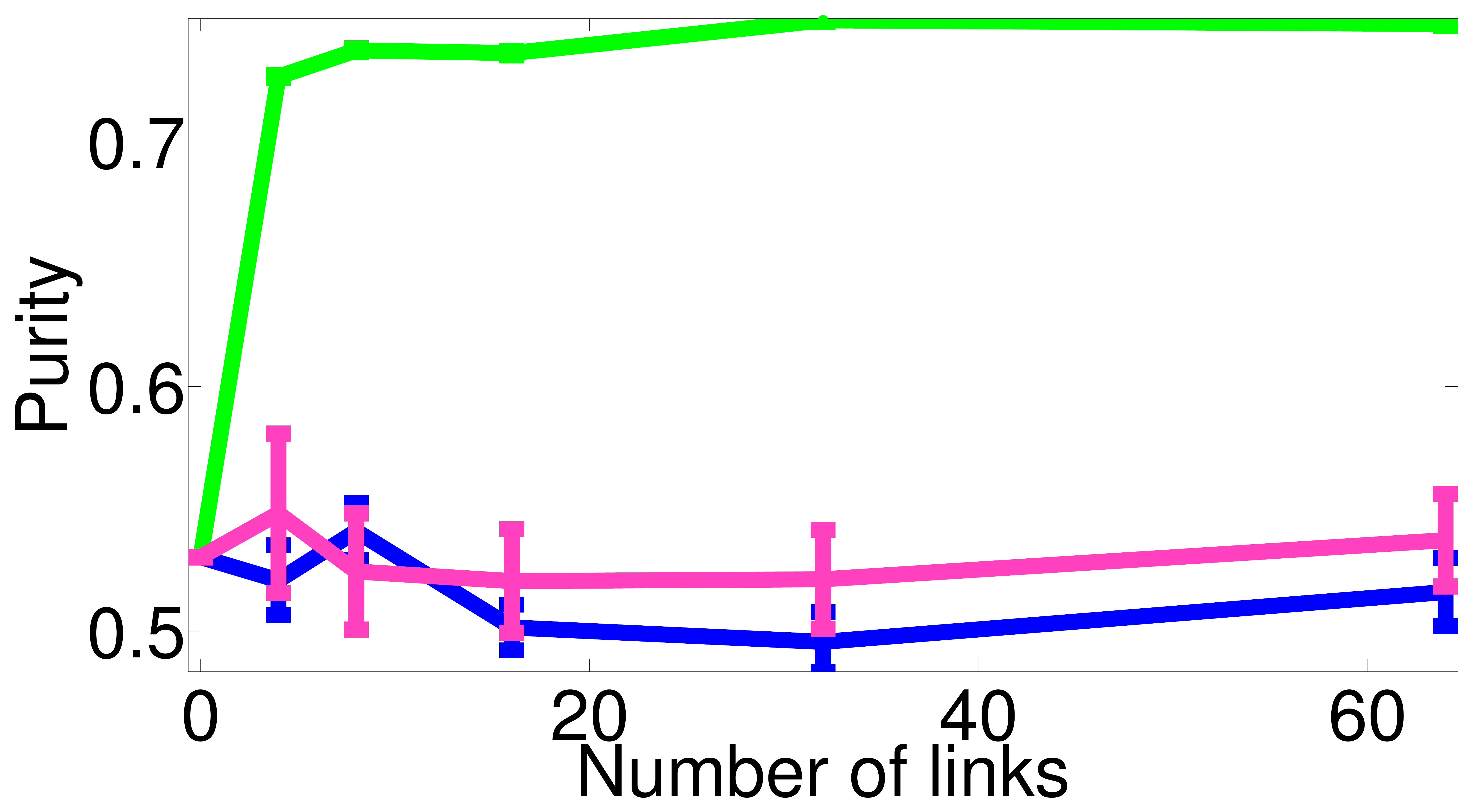}
		{(c) digit 1, 2, and 3}{(d) digit 4, 5, 6, and 7}
	\end{minipage}
	\vspace{-5pt}
	\caption
	[Result of MNIST and UCI]
	{
	The performance of {\bf GM-PR} compared to 
	{\bf GMM-EC}~\cite{shental2004computing} 
	with a different number of pairwise links on 
	(a) Harberman, (b) Thyroid, (c) digits 1, 2, and 3, and (d) digits 4, 5, 6, and 7.
	}
 	\label{fig:GMMECBOTH}
	\vspace{-5pt}
\end{figure*}
\begin{figure*}
	\twoAcrossLabels{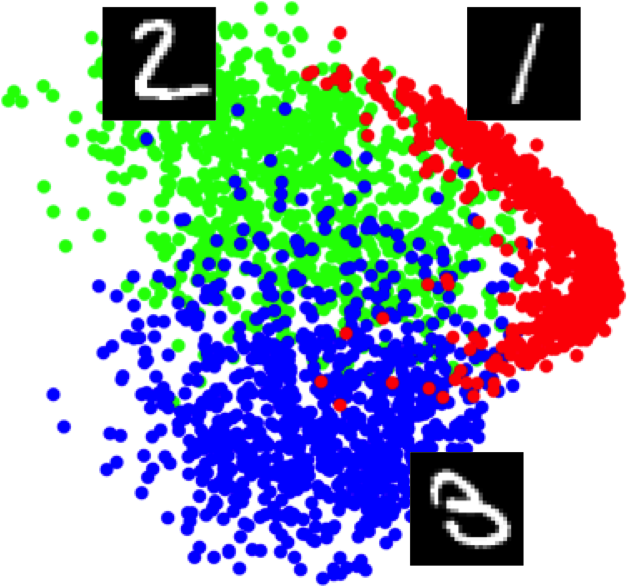}{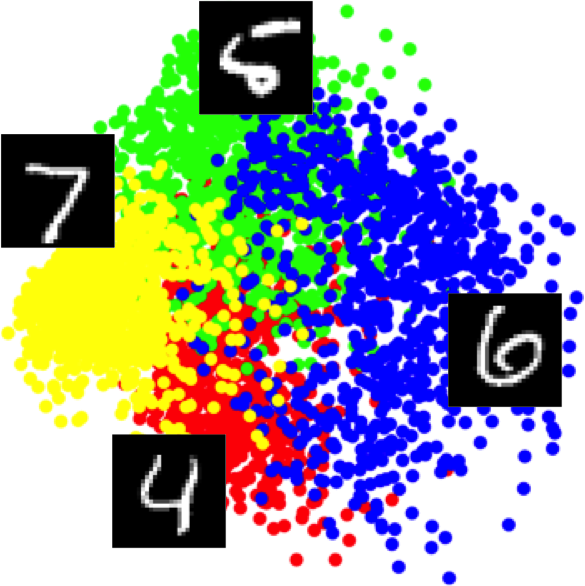}
	{(a) digits 1, 2, and 3}{(b) digits 4, 5, 6, and 7}
	\centering
  	\caption
	[Visualization of MNIST]
	{
		Digits 1, 2, and 3, and digits 4, 5, 6, and 7 visualized by the first two principal components of PCA. 
	}
 	\label{fig:MNISTEX}
\end{figure*}
\begin{figure*}[!t]
	\center
	\begin{minipage}[b]{0.1\linewidth}
		\includegraphics[width=2.1cm,height=1.4cm]{./fig/GMMEClegend-eps-converted-to.pdf}
	\end{minipage}%
	\begin{minipage}[t]{\linewidth}
		\twoAcrossLabels{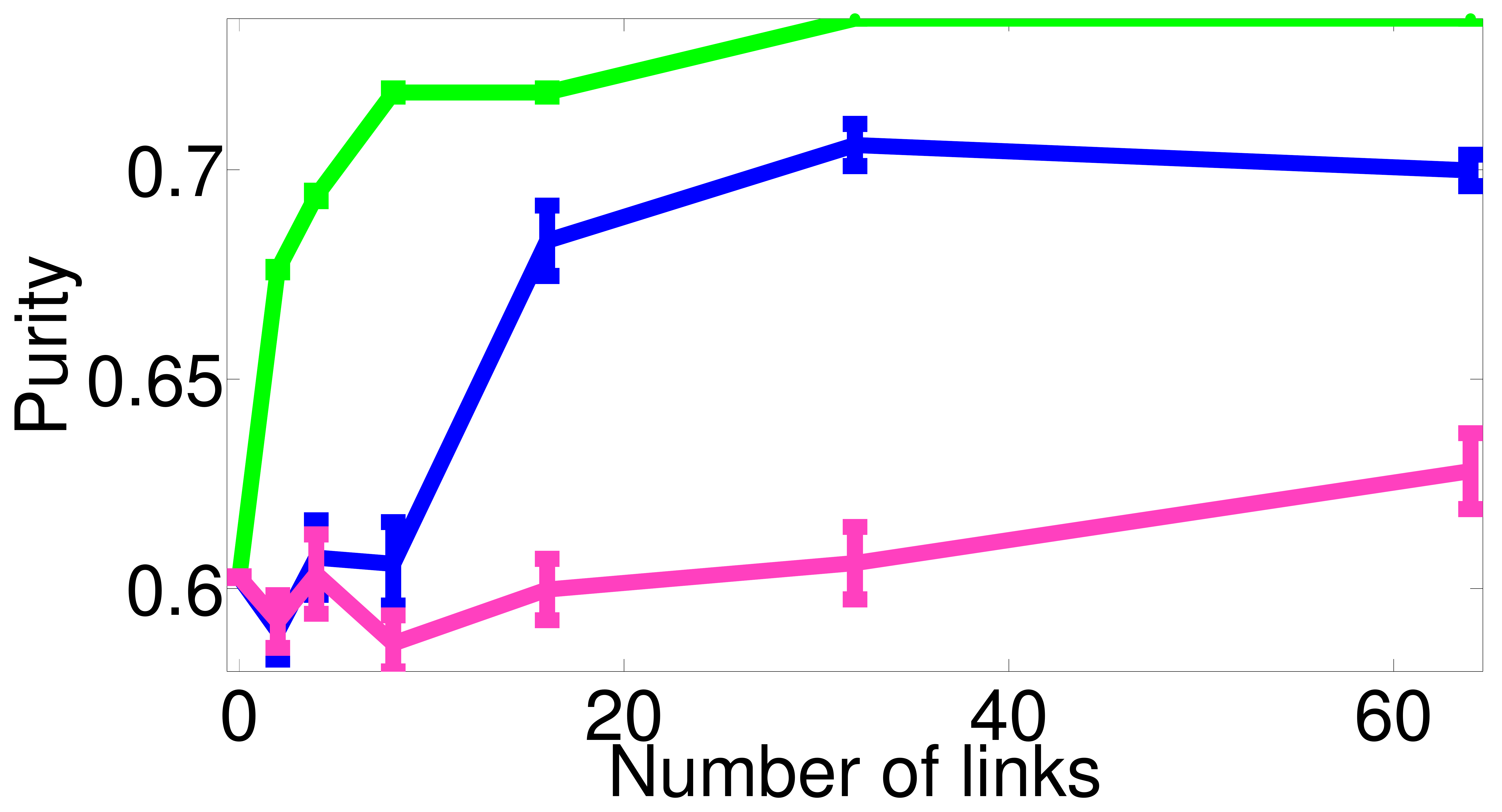}{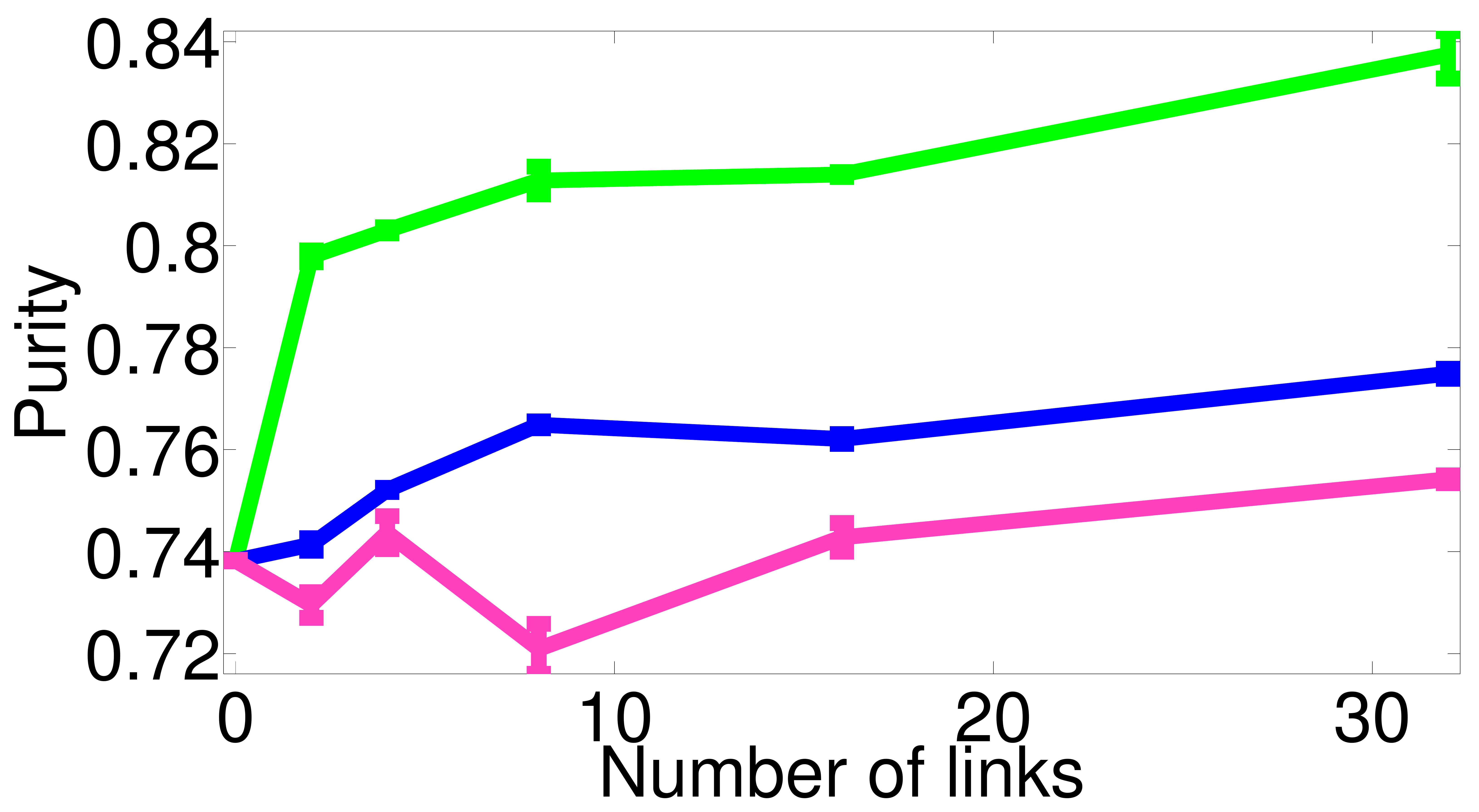}
		{(a) Harberman used only must-links}{(b) Thyroid used only must-links}
		\twoAcrossLabels{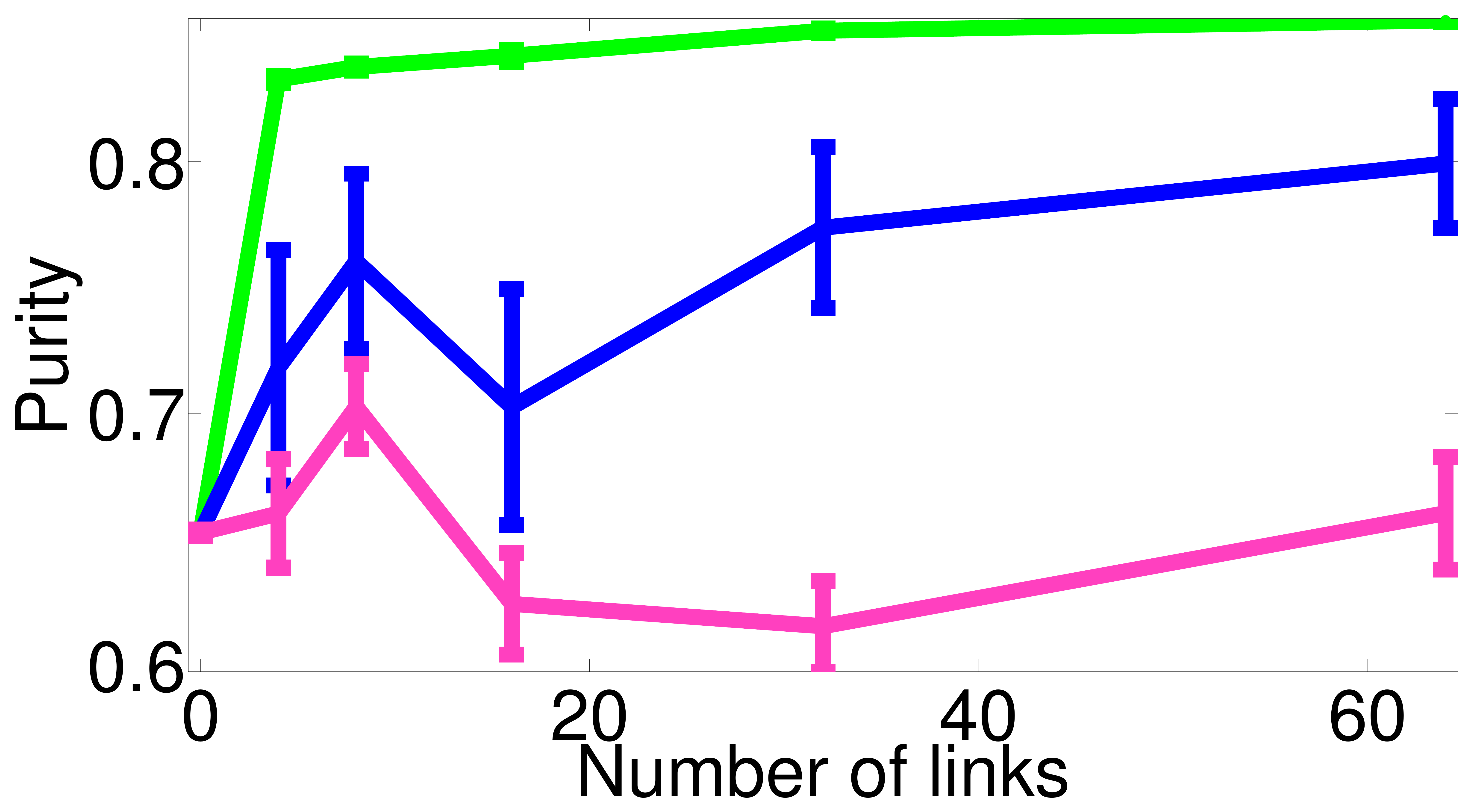}{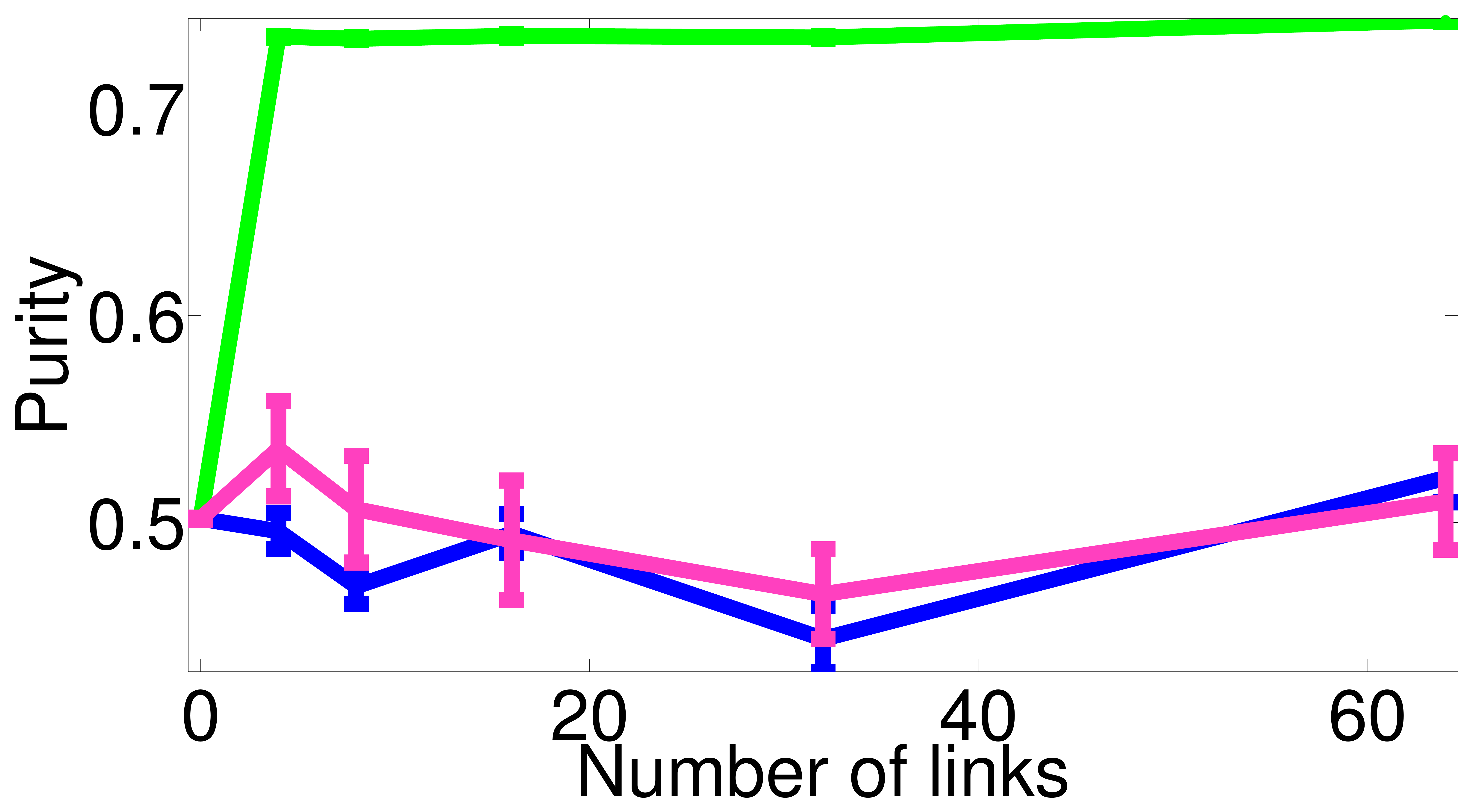}
		{(c) digits 1, 2, and 3}{(d) digits 4, 5, 6, and 7}
	\end{minipage}
	\vspace{-5pt}
	\caption
	[Result of MNIST and UCI with only must-link relations]
	{
	The performance of {\bf GM-PR} compared to 
	{\bf GMM-EC}~\cite{shental2004computing} 
	with a different number of must-links on 	
	(a) Harberman, (b) Thyroid, (c) digits 1, 2, and 3, and (d) digits 4, 5, 6, and 7.
	}
 	\label{fig:GMMECML}
	\vspace{-5pt}
\end{figure*}
\begin{figure*}[!t]
	\center
	\begin{minipage}[b]{0.1\linewidth}
		\includegraphics[width=2cm,height=1.4cm]{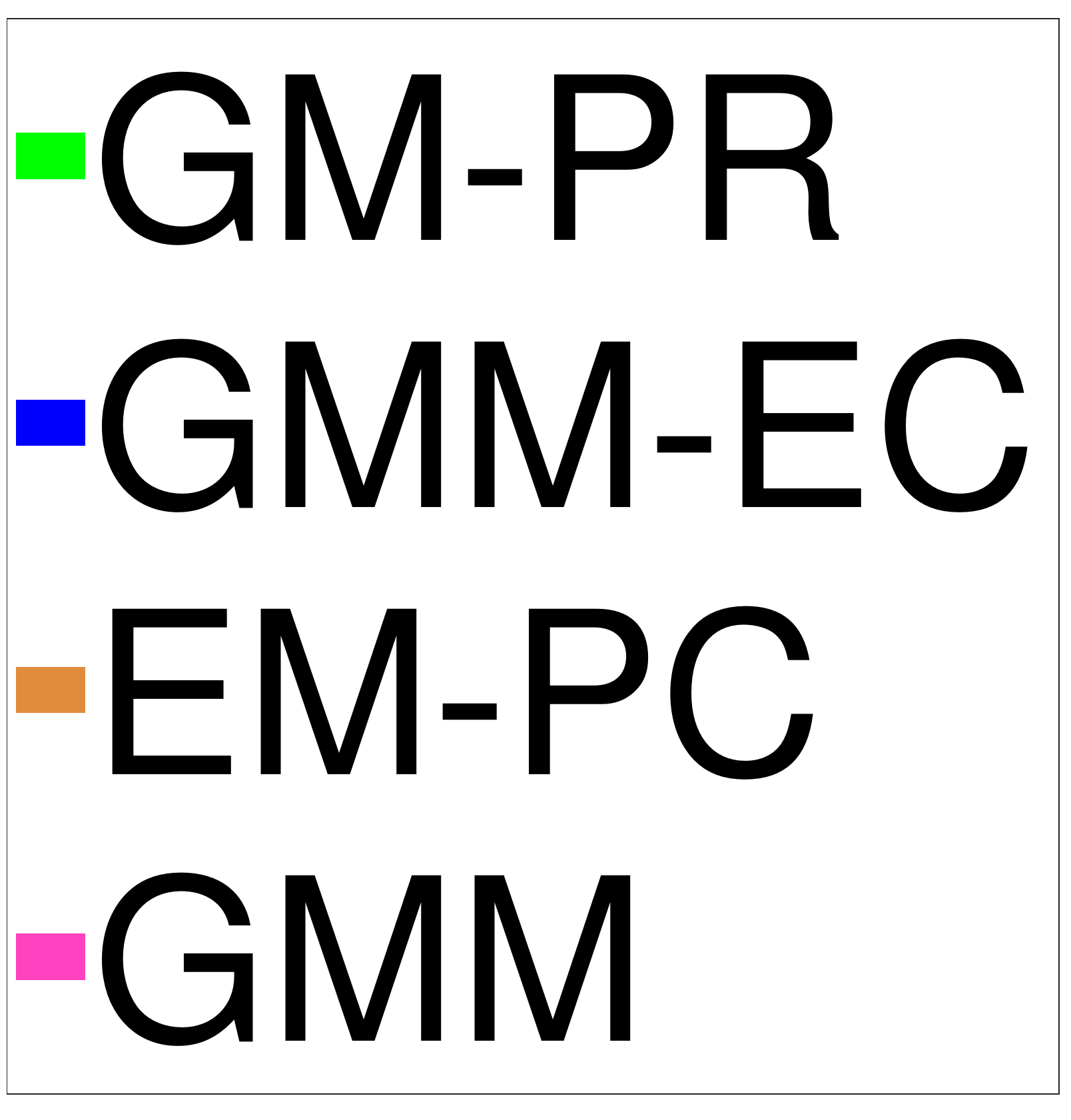}
	\end{minipage}%
	\begin{minipage}[t]{\linewidth}
		\twoAcrossLabels{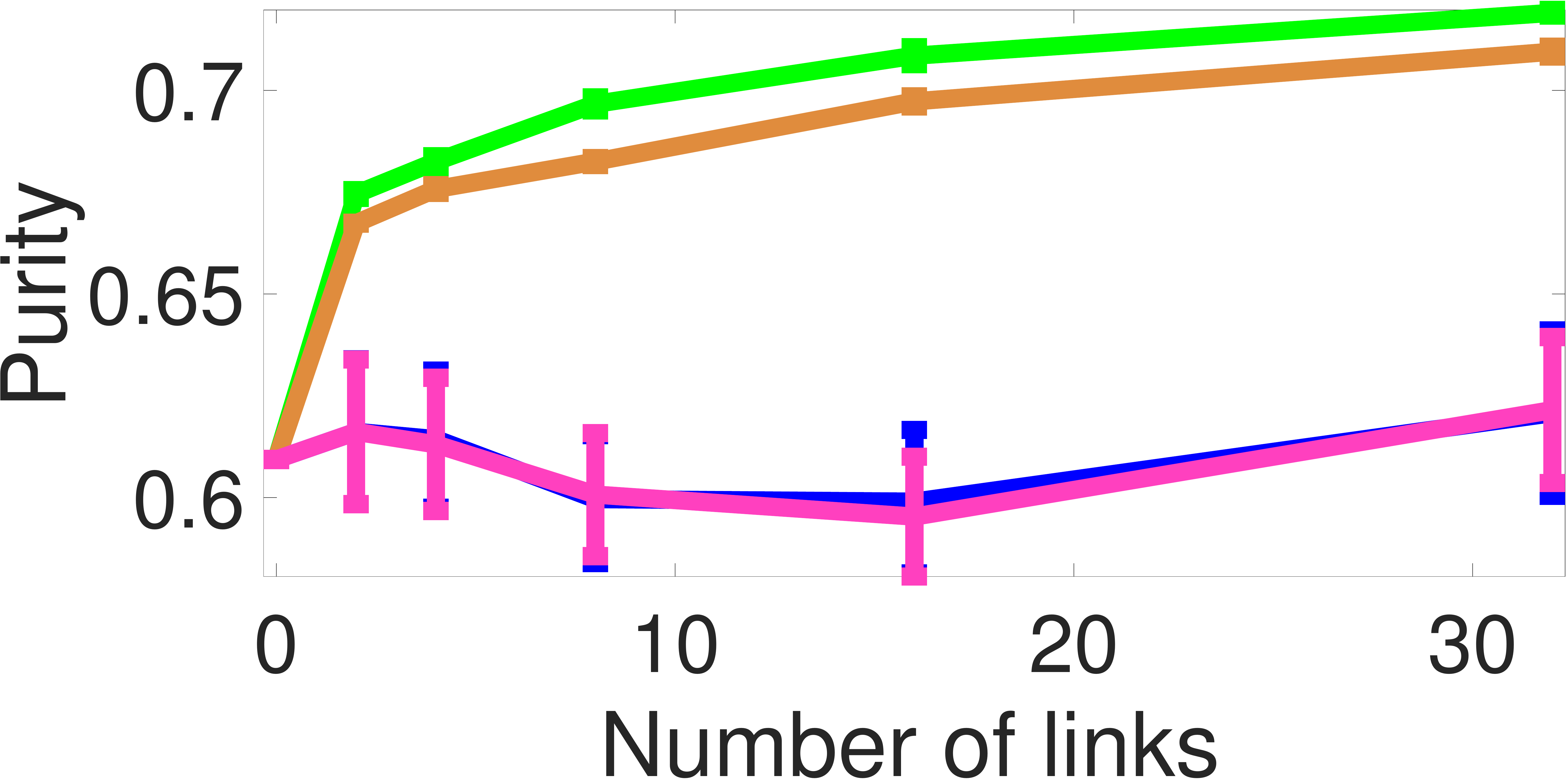}{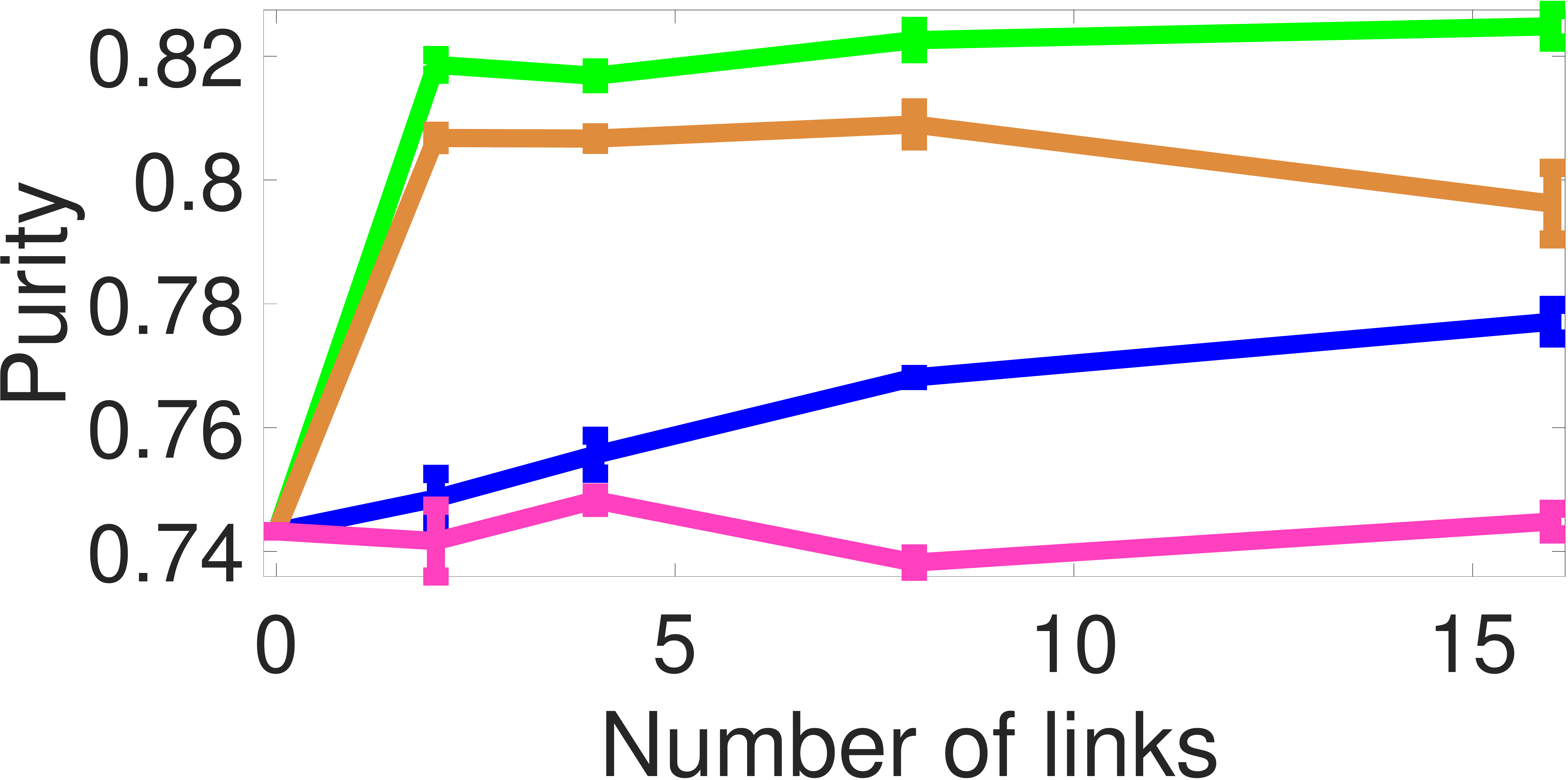}
		{(a) Harberman}{(b) Thyroid}
		\twoAcrossLabels{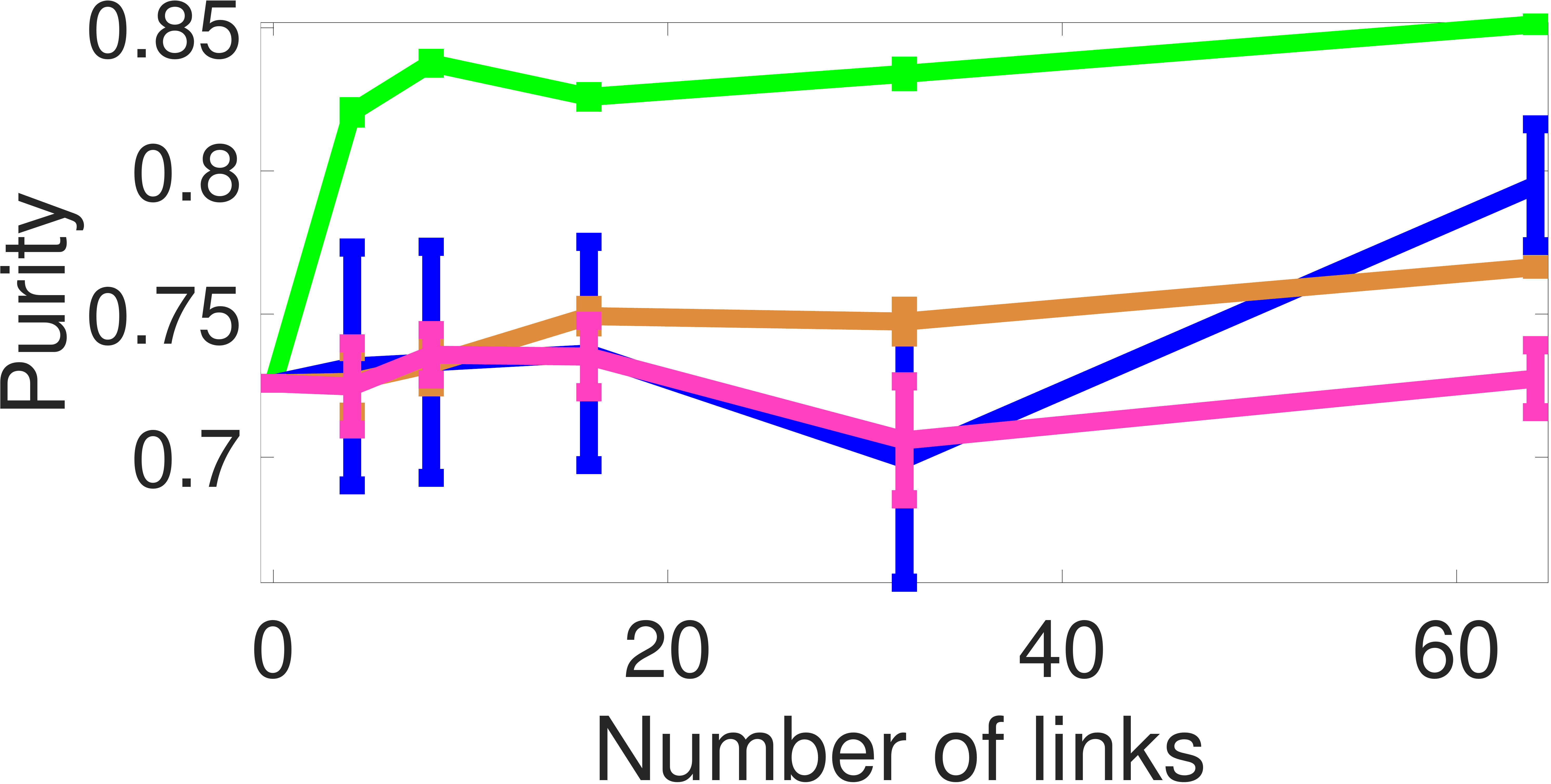}{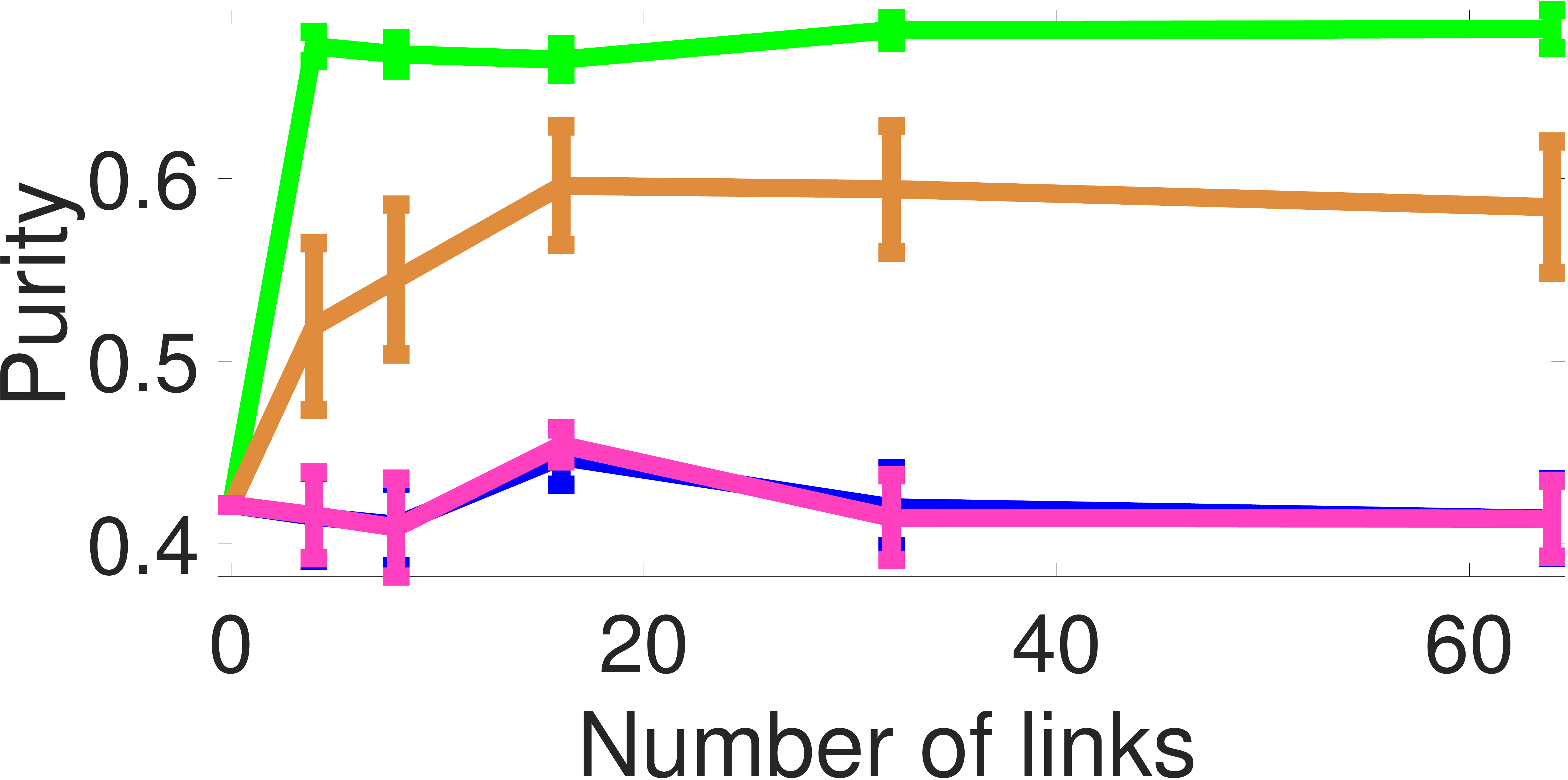}
		{(c) digits 1, 2, and 3}{(d) digits 4, 5, 6, and 7}
	\end{minipage}
	\vspace{-5pt}
	\caption
	[Result of MNIST and UCI with only cannot-link relations]
	{
	The performance of {\bf GM-PR} compared to 
	{\bf GMM-EC}~\cite{shental2004computing} and 
	{\bf EM-PC}~\cite{conf/nips/GracaGT07} 
	with a different number of cannot-links on
	(a) Harberman, (b) Thyroid, (c) digits 1, 2, and 3, and (d) digits 4, 5, 6, and 7.
	}
 	\label{fig:EMPC}
	\vspace{-5pt}
\end{figure*}

We demonstrate the performance of {\bf GM-PR} 
on two binary clustering tasks, Haberman and Thyroid, 
and two multiclass problems, digits 1, 2, 3 and 4, 5, 6, 7.
For ease of visualization, 
we work with only the leading two principal components of the MNIST 
using principal component analysis (PCA).
Figure~\ref{fig:MNISTEX} shows two-dimensional inputs, color-coded by class label.
Figure~\ref{fig:GMMECBOTH} shows that 
{\bf GM-PR} significantly outperforms {\bf GMM-EC} 
regardless of the available number of links on all datasets.
Moreover, 
Figure~\ref{fig:GMMECML} shows that {\bf GM-PR} performs well 
even if only the {\bf must-links} are available.
Compared to {\bf EM-PC}, which uses only the {\em cannot-links}, 
Figure~\ref{fig:EMPC} shows the performance of {\bf GM-PR}
is always greater than or comparable to {\bf EM-PC} and {\bf GM-PR}. 
Figure~\ref{fig:EMPC} also shows that the performance of {\bf EM-PC} decreases 
when the number of classes increases.
The cannot-link in the {\bf GM-PR}, on the other hand, can contribute to the model when the problem is either binary or multiclass.
Notice that all the experiments indicate that {\bf GM-PR} has a lower variance over 100 random initializations, which implies {\bf GM-PR} stability regardless of the number of available pairwise links.

\subsection{Results: Mixture of Gaussians ($\mK>1$)}
In this section, we demonstrate the performance of the proposed model using a
mixture of Gaussians on the datasets that have local manifold structure.

\label{subsec:MG}
\subsubsection{Synthetic Data: Two Moons Dataset}
\label{subsec:MM:twoMoon}
\begin{figure*}[!t]
	\twoAcrossLabels{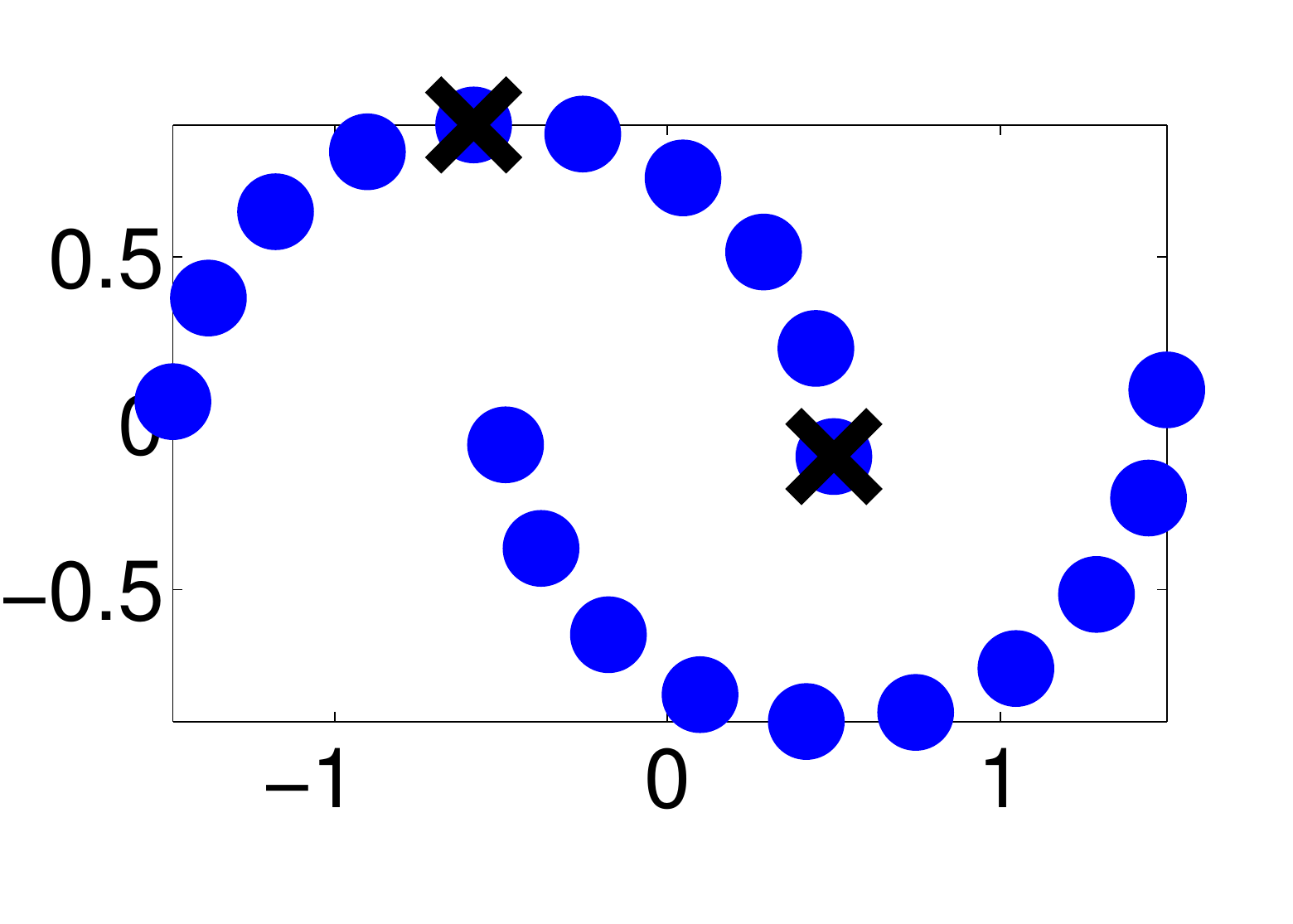}{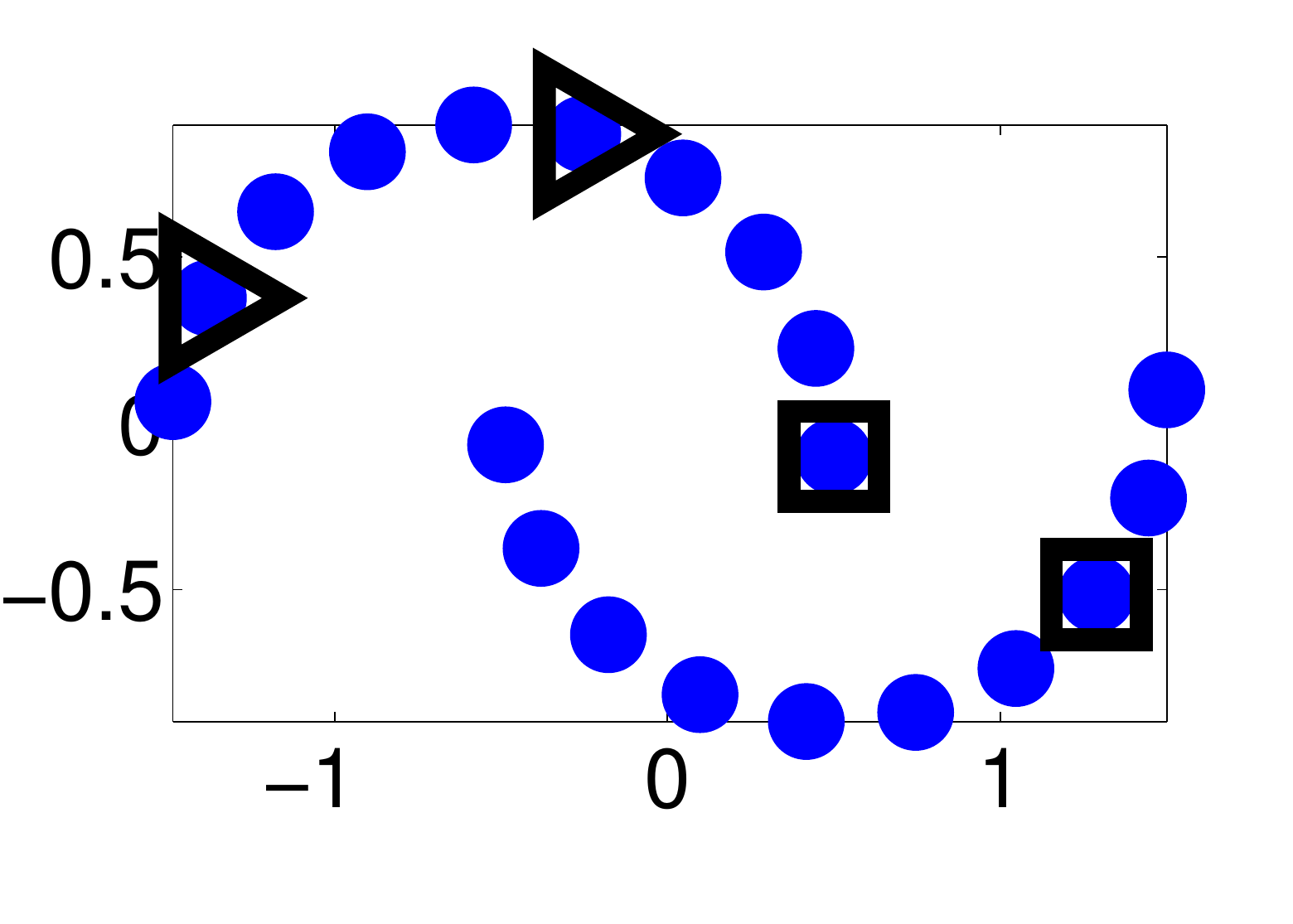}
	{(a)}{(b)}
	\threeAcrossLabels{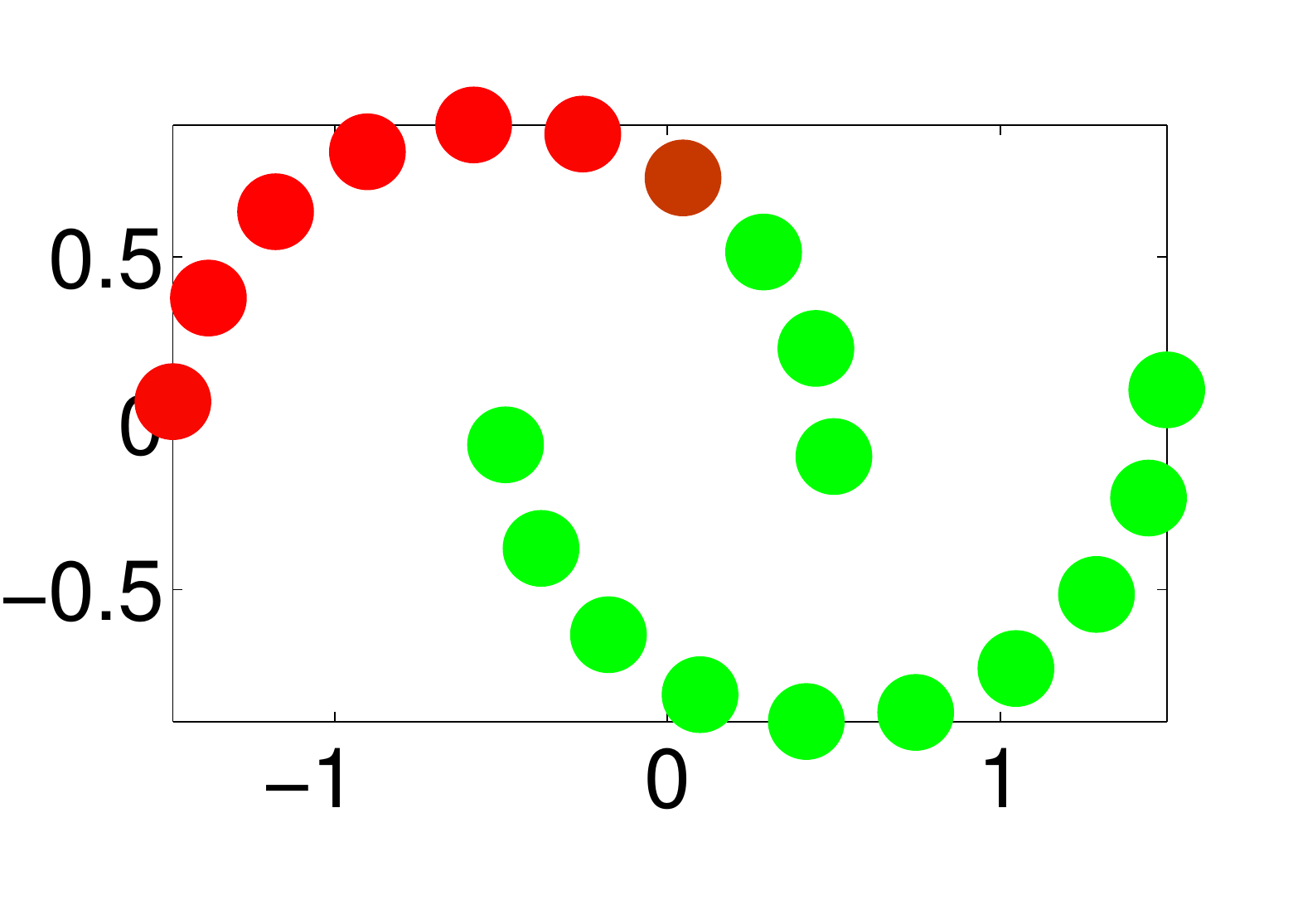}{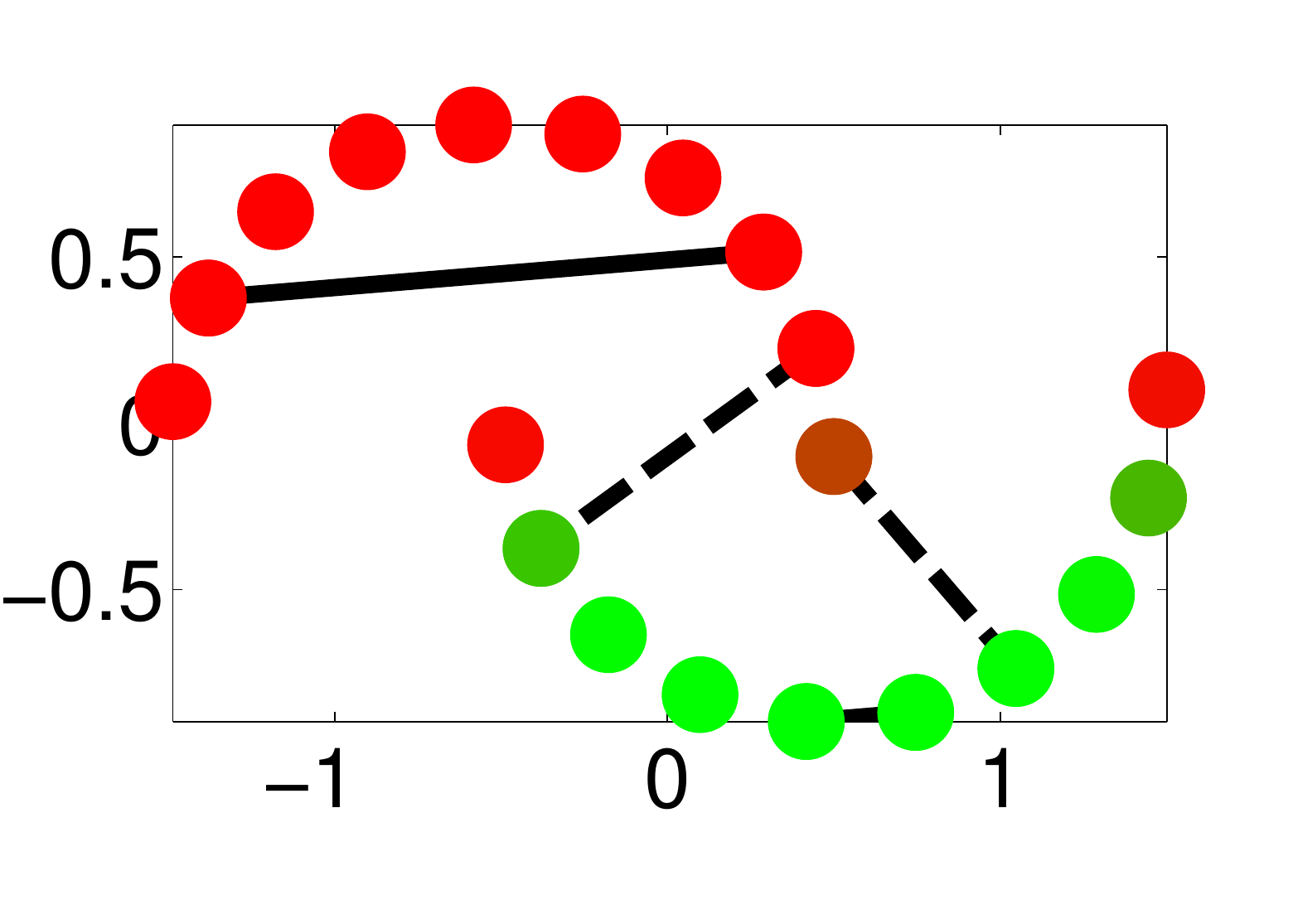}{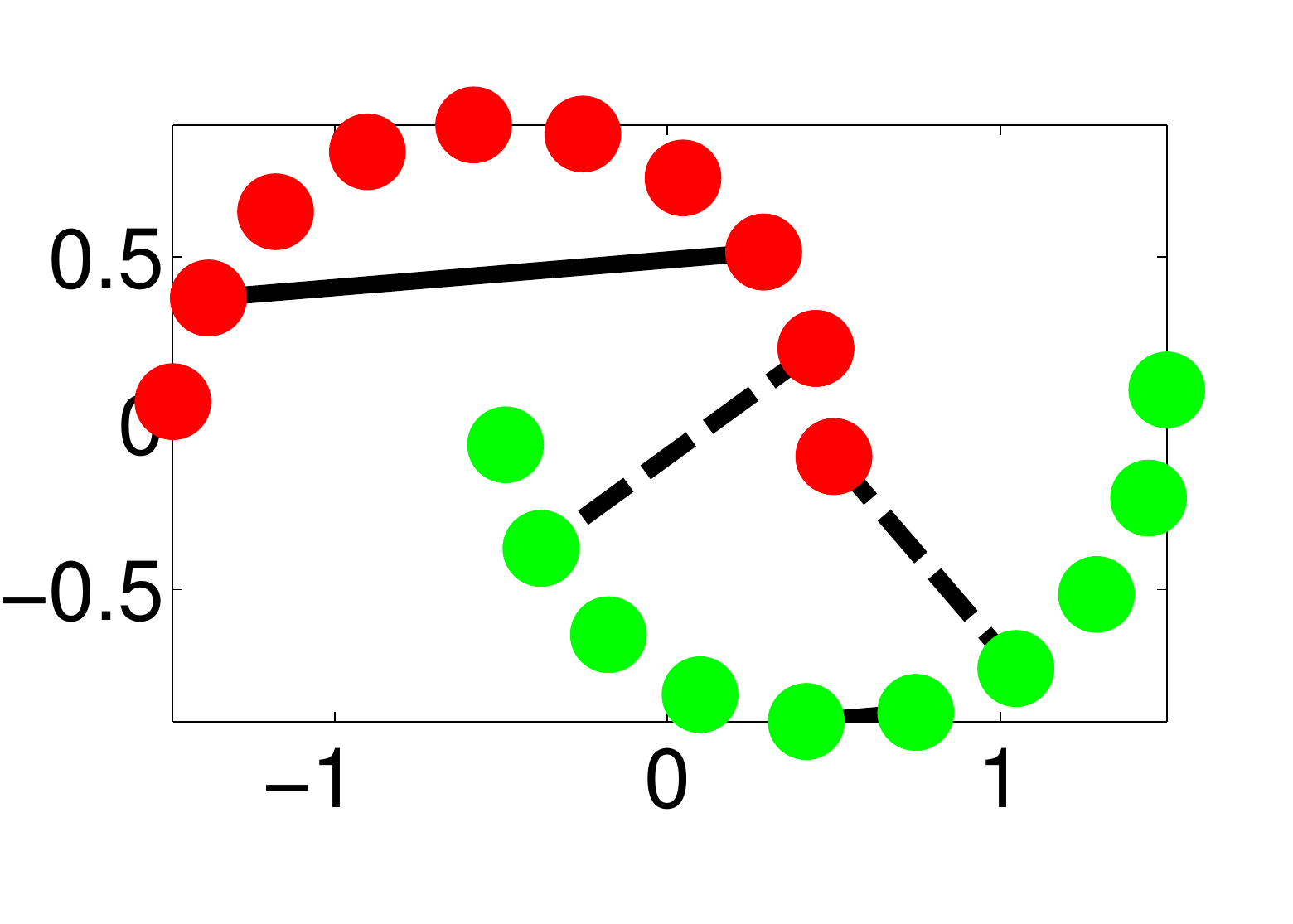}
	{(c)}{(d)}{(e)}
     \vspace{-5pt}
	\caption
	[Result of two moons synthetic data]
	{ {\bf  Two moons synthetic data}: 
	(a) Original data with initial two means marked by x.
	(b) Original data with initial means marked by triangles for class 1 and squares for class 2.
	Results are represented as follows:
	(c) {\bf GMM},
	(d) {\bf GM-PR} used one cluster for each class, 
	and two {\em must-links} (solid line)
	and two {\em cannot-links} (dashed line), and 
	(e) {\bf GM-PR} used two clusters for each class and used the same links as in (d).
	}
 	\label{fig:twomoons}
	\vspace{-5pt}
\end{figure*}

Data points in two moons are on a moon-like manifold structure (Figure~\ref{fig:twomoons}(a)),
which allows us to show the advantage of the proposed method
using a mixture of Gaussians as a distribution instead of a single Gaussian distribution.
Figure~\ref{fig:twomoons}(a) shows the data with initial means 
for the {\bf GMM} and the {\bf GM-PR} using a single Gaussian.
Figure~\ref{fig:twomoons}(b) shows the data with initial means
for {\bf GM-PR} using a mixture of Gaussians ($\mK=2$).
Figure~\ref{fig:twomoons}(c) is the clustering result obtained 
from the unconstrained {\bf GMM},
in which
three points were assigned to the wrong class. 
Figure~\ref{fig:twomoons}(c) also shows that the performance of the {\bf GMM} relied on the parameter initialization.
Figure~\ref{fig:twomoons}(d) shows that the proposed \textbf{GM-PR}, 
which used a single cluster for each class,
tried to learn the manifold structure
via two must-link and two cannot-link relations. 
However, two points were still assigned to the incorrect class.
Figure~\ref{fig:twomoons}(e) shows that the \textbf{GM-PR} 
can trace the manifold structure but used the same links in (d) 
with two clusters for each class.
This experiment illustrates the advantage of the proposed model 
with a mixture of distributions 
that traces the local data structure by every single cluster 
and describes the global data structure using the mixture of clusters.

\subsubsection{COIL 20}
\label{subsubsec:MM:RealData}
\begin{figure*}[!t]
	\center
	\begin{minipage}[b]{0.1\linewidth}
		\includegraphics[width=1.25cm,height=1.25cm]{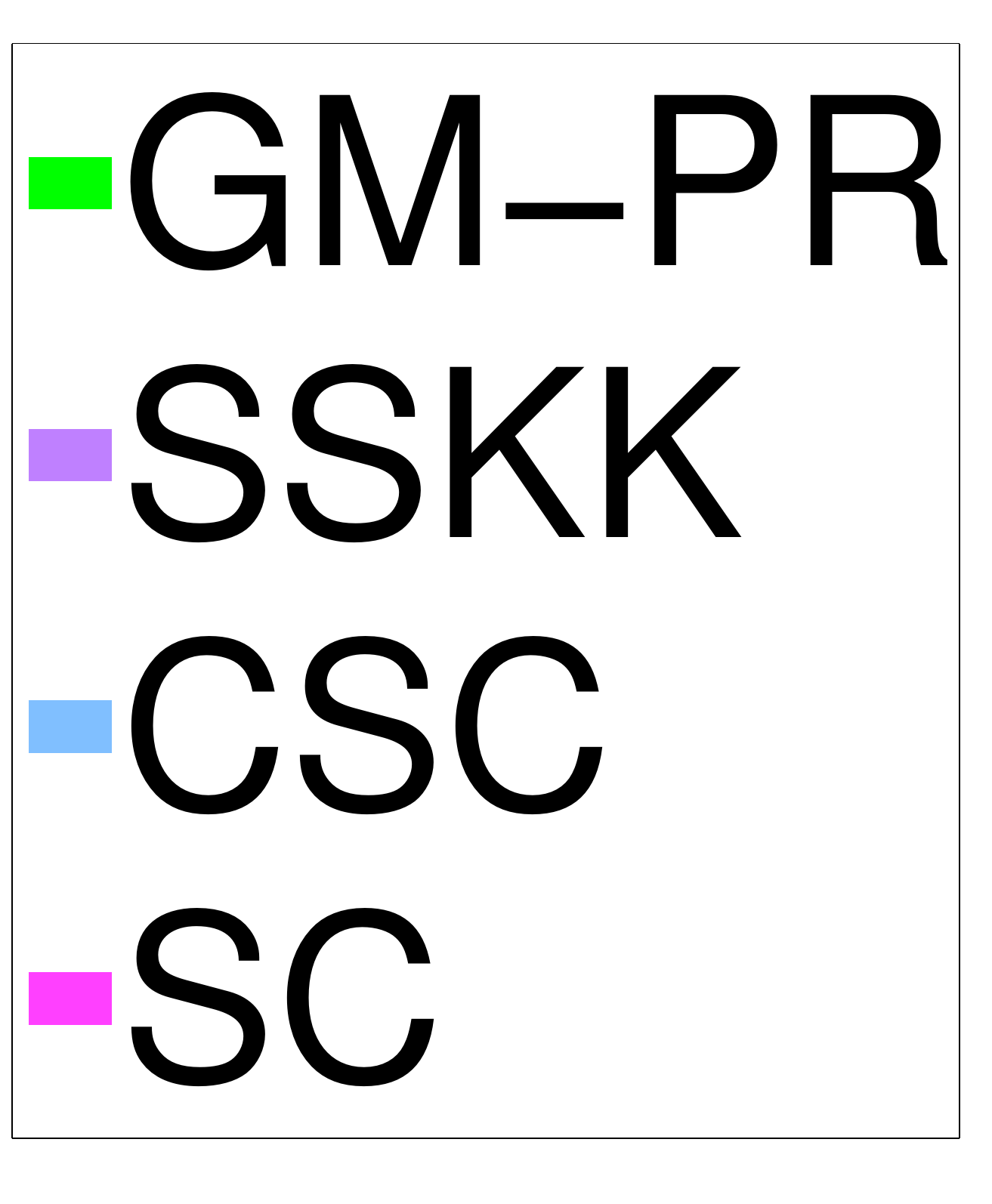}
	\end{minipage}%
	\begin{minipage}[t]{\linewidth}
		\twoAcrossLabels{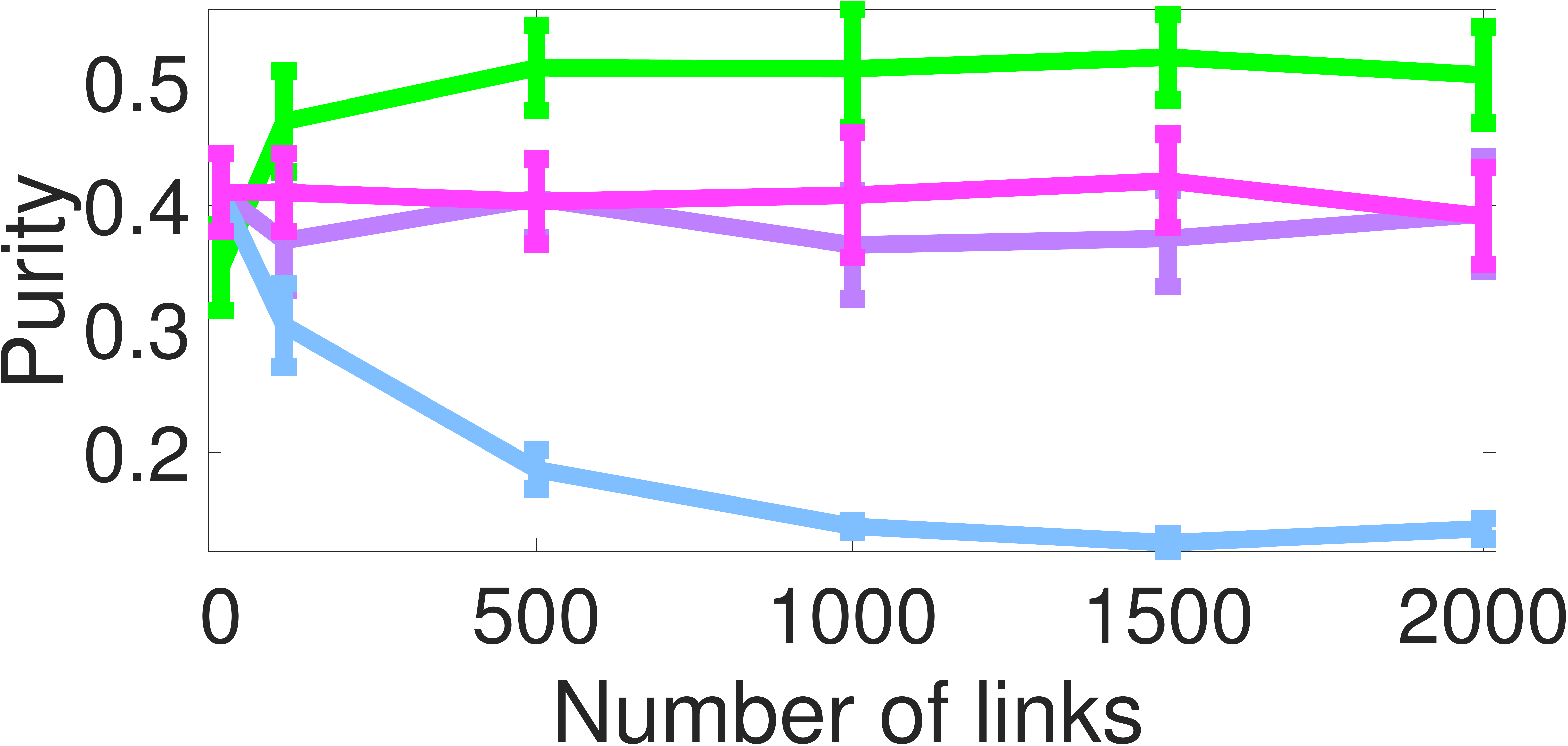}{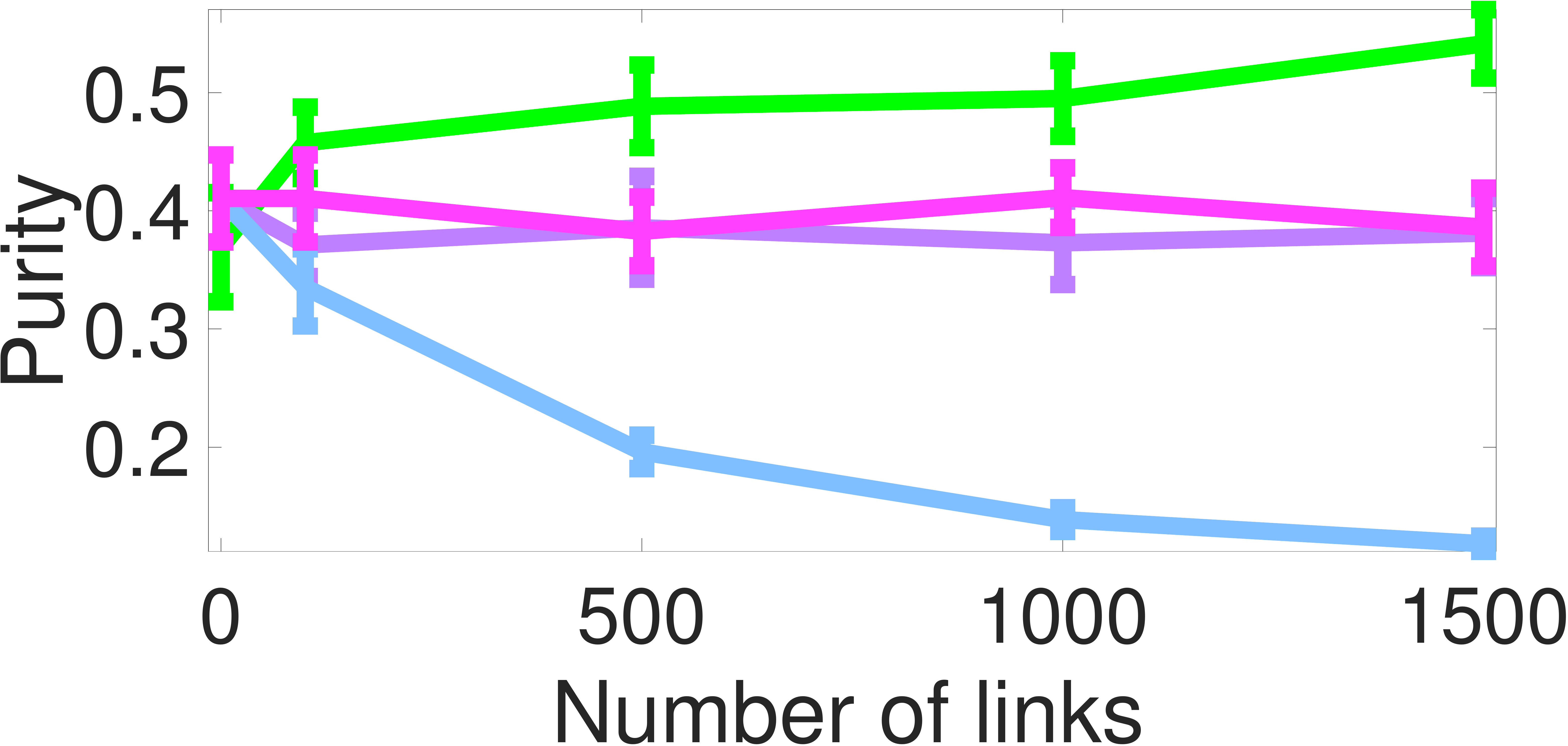}
		{(a) d = 10}{(b) d = 15}
		\twoAcrossLabels{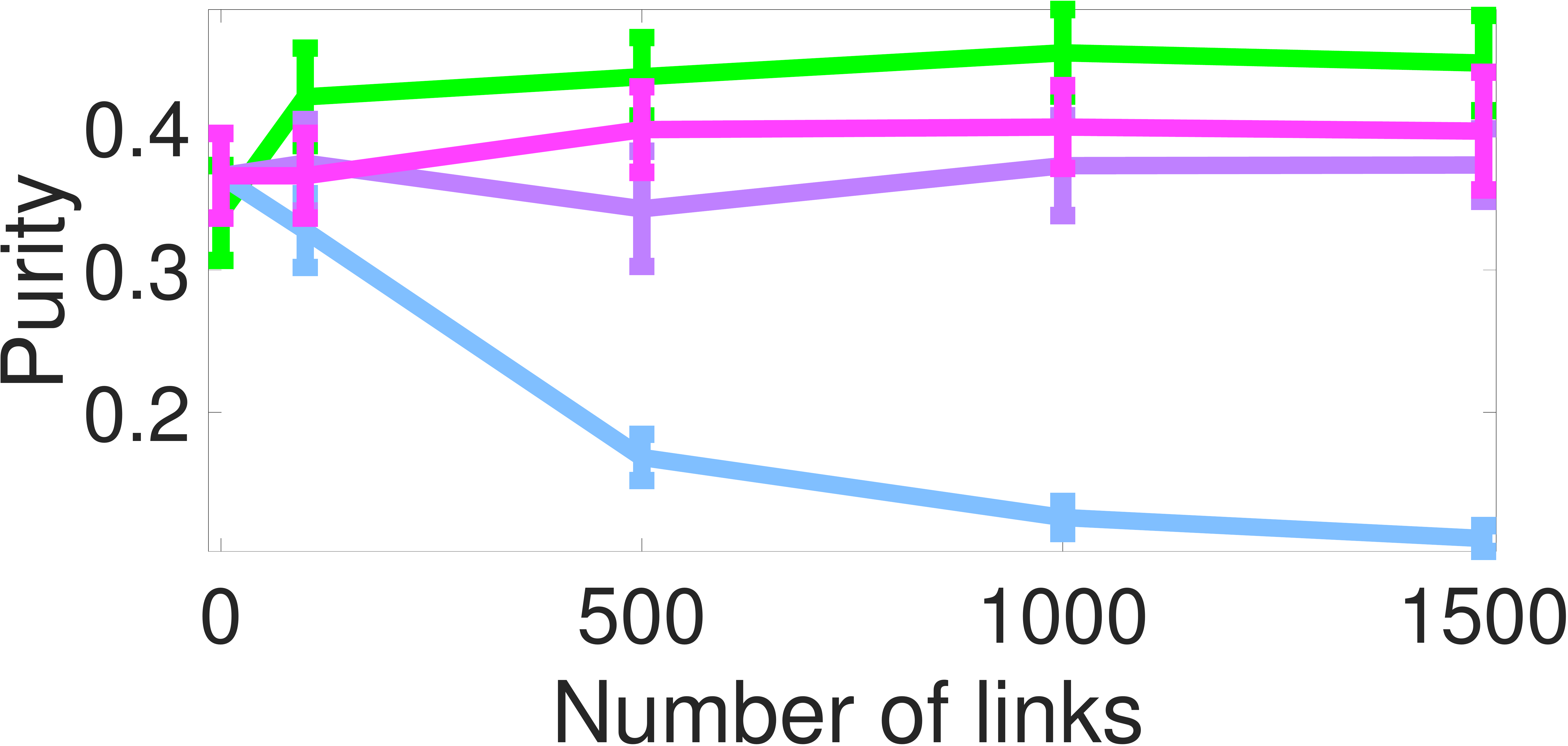}{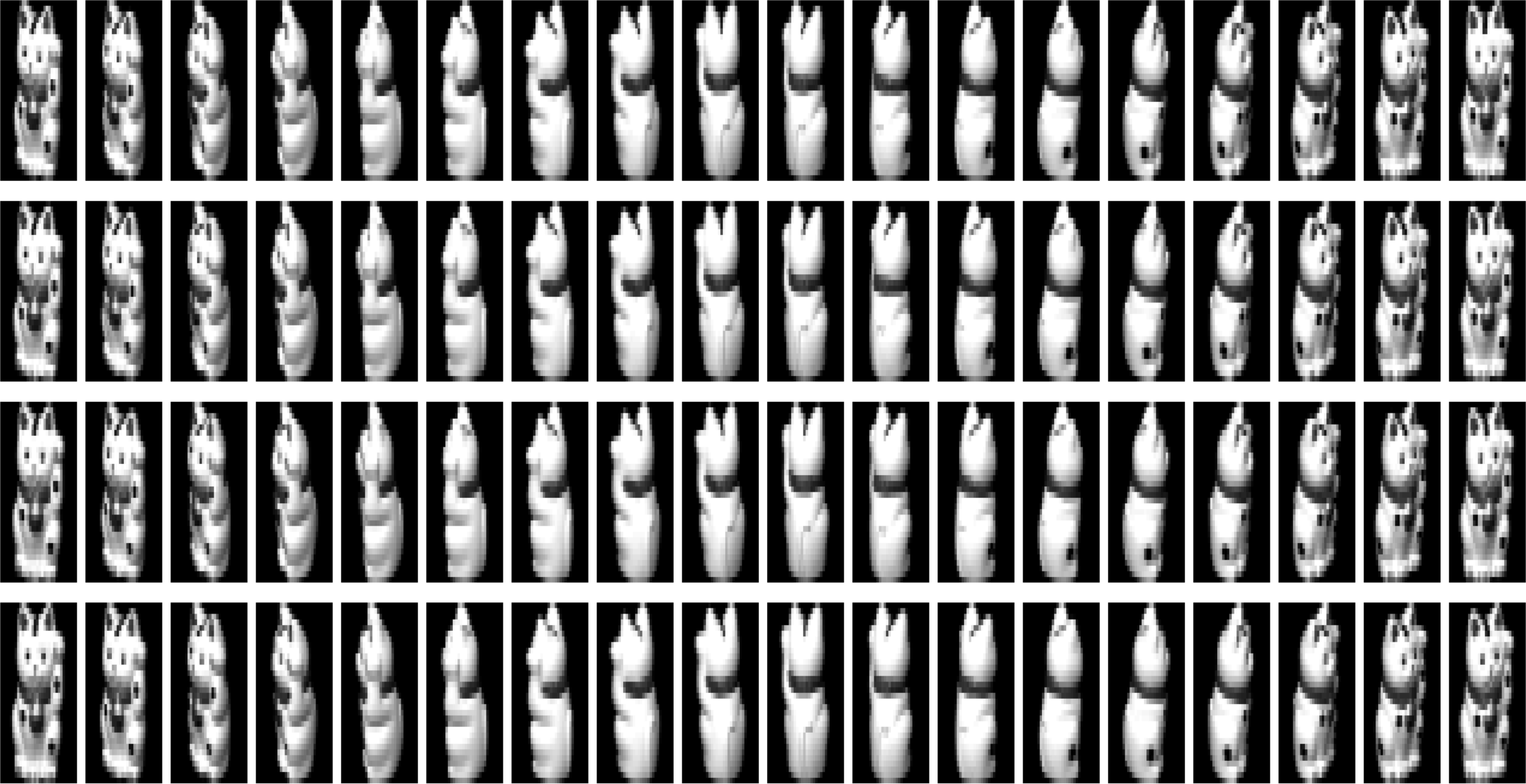}
		{(c) d = 20}{ COIL-20}
	\end{minipage}
	\vspace{-5pt}
	\caption
	[Result of COIL 20]
	{
		The performance of {\bf GM-PR} compared to 
		{\bf SSKK}~\cite{kulis2009semi} and 
		{\bf CSC}~\cite{wang2010flexible} 
		with a different number of cannot-links on COIL-20 (d),
		which are projected onto the low-dimensional subspace: 
		(a) d = 10 (b) d = 15, and (c) d = 20.
	}
 	\label{fig:COIL20}
	\vspace{-5pt}
\end{figure*}

In this section, we report the performance of
COIL 20\footnote{http://www.cs.columbia.edu/CAVE/software/softlib/coil-20.php} 
datasets, which contain images of 20 objects in which 
each object was placed on a turntable and rotated 360 degrees to be captured 
with different poses via a fixed camera~(Figure~{\ref{fig:COIL20}}). 
The COIL 20 dataset contains 1440 instances and 1024 attributes.
We set the number of multiclusters per class by cross-validation to $\mK=3$.
Previous studies have shown that the intrinsic dimension of many high-dimensional real-world datasets is often quite small ($d \leq 20$)~\cite{raginsky2005estimation,felsberg2009continuous};
therefore, each image is first projected onto the low-dimensional subspace (d = 10, 15, and 20).
Figure~{\ref{fig:COIL20}} shows that the {\bf GM-PR} provides higher purity values 
compared to the {\bf SSKK} and the {\bf CSC} with fewer links ($\leq 1000$) 
regardless of the subspace dimension.
In these experiments, 
we found that the proposed model can outperform the graph-based method with fewer links.

\subsection{Result: Sensitivity to Number of Clusters Per Class}
\label{subsubsec:MM:RealData}
\begin{figure*}[!t]
	\centering
	\begin{minipage}[b]{0.1\linewidth}
		\includegraphics[width=1cm,height=1.5cm]{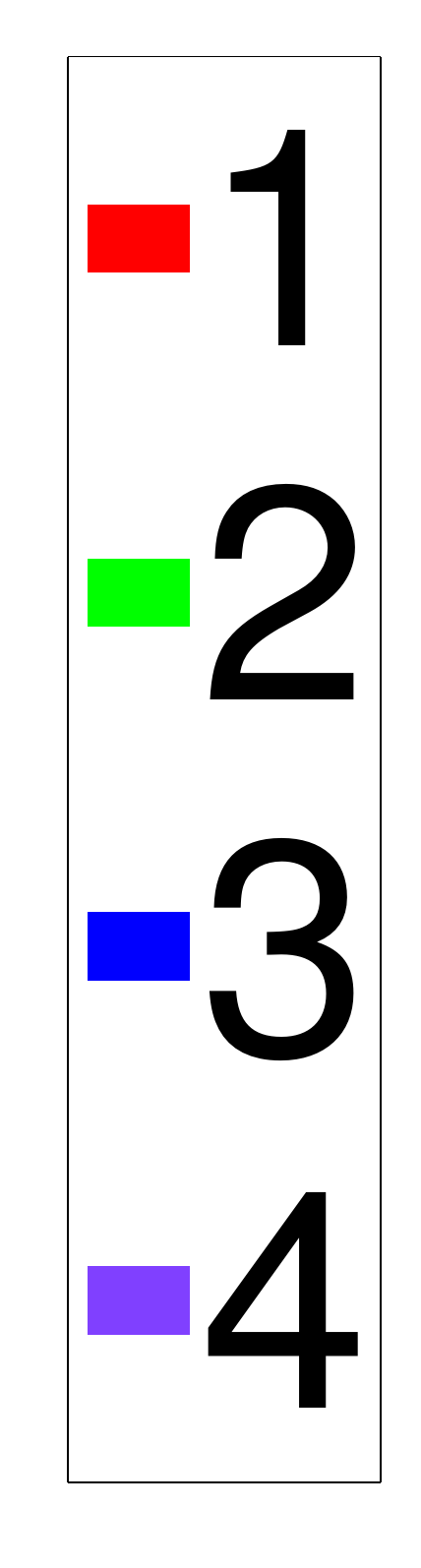}
	\end{minipage}%
	\begin{minipage}[t]{\linewidth}
		\twoAcrossLabels{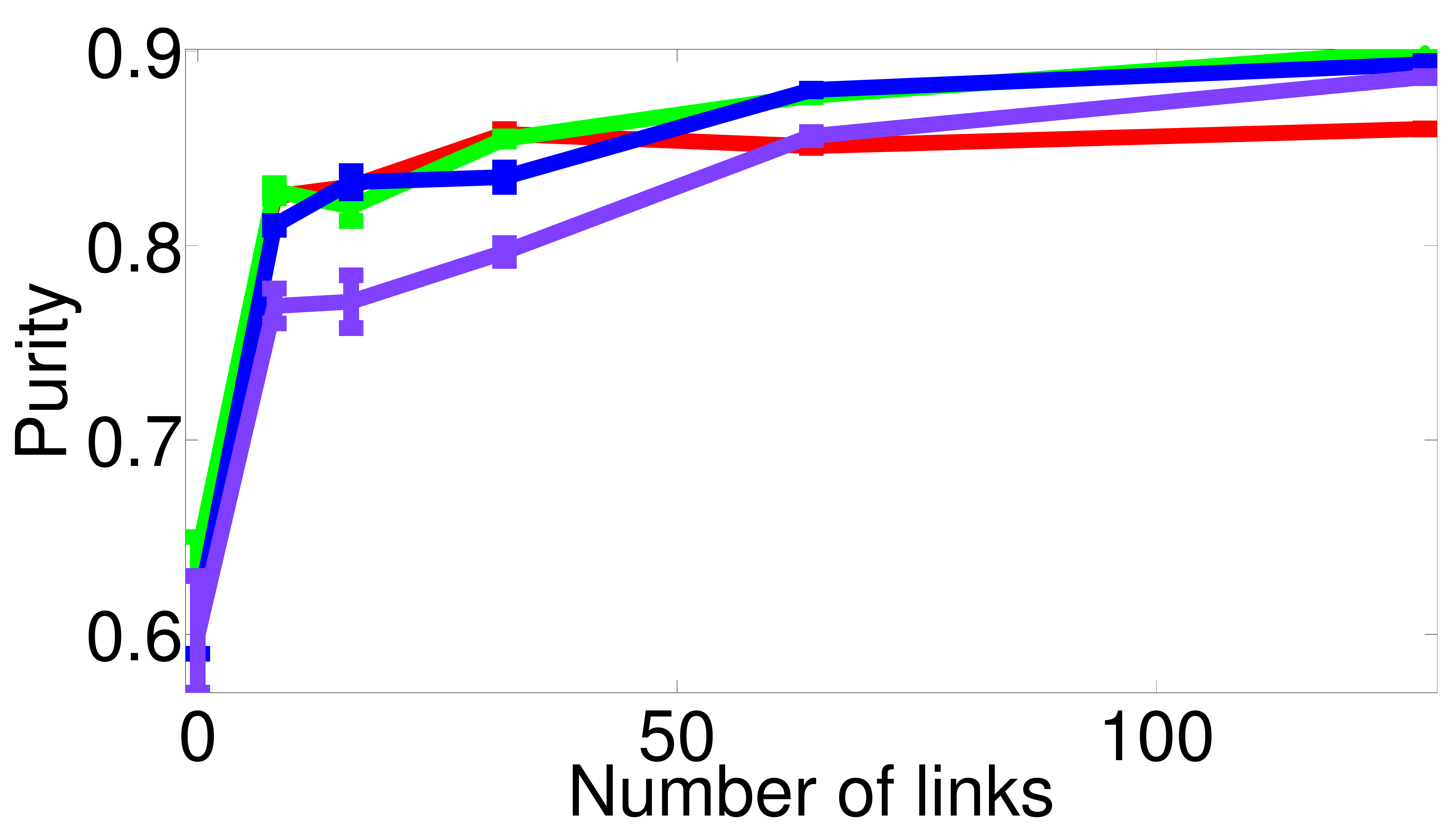}{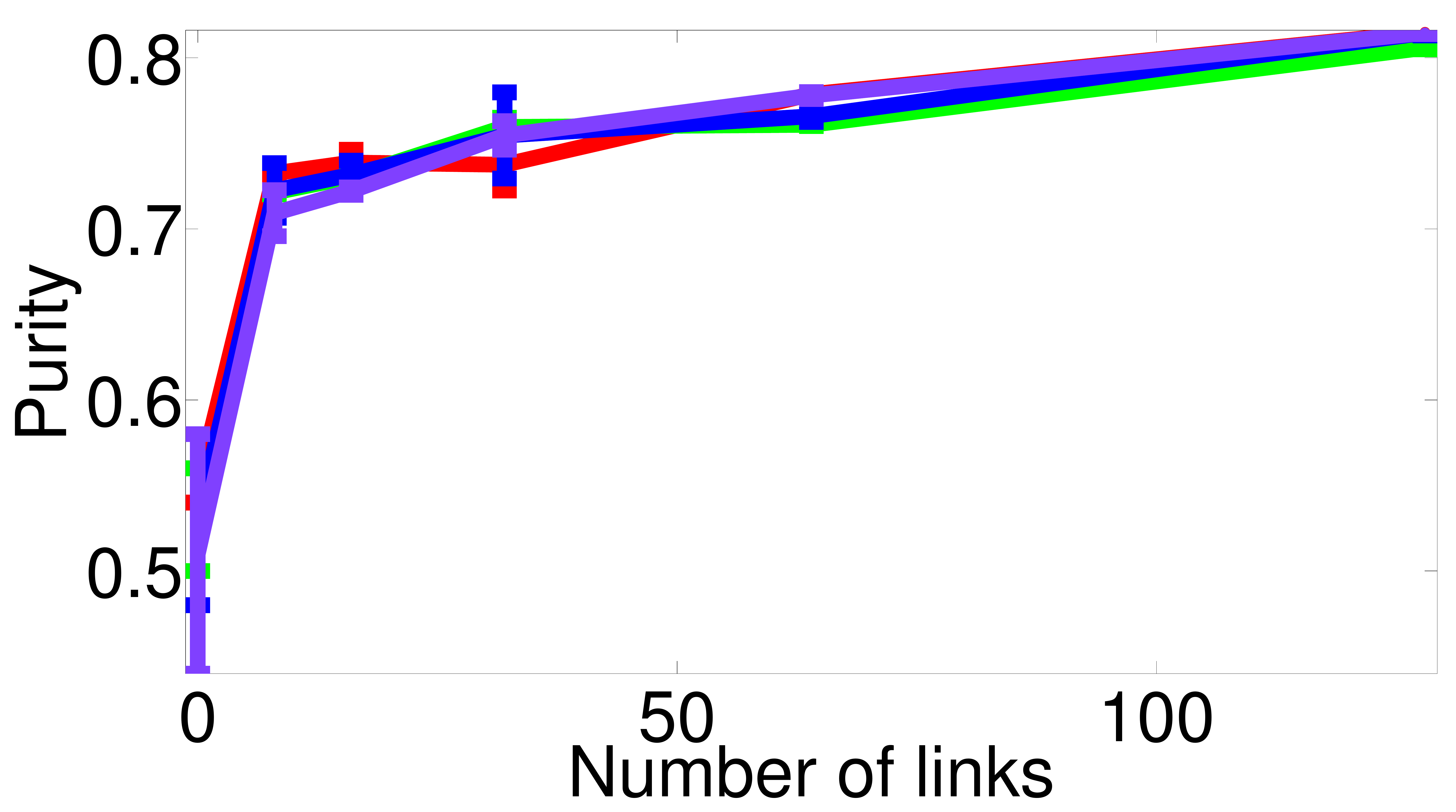}
		{(a) digits 1, 2, and 3}{(b) digits 4, 5, 6, and 7}
	\end{minipage}
	\vspace{-5pt}
	\caption
	{
		The performance of {\bf GM-PR} uses different values of $\mK$ on 
		(a) digits 1, 2, and 3 and (b) digits 4, 5, 6, and 7.
	}
 	\label{fig:MK}
	\vspace{-5pt}
\end{figure*}

Lastly, we demonstrated the performance of the proposed model 
in regard to different values of $\mK$. 
First, we used the same dataset (MNIST) that is used in section~\ref{subsubsec:G:RealData}.
In Figure~\ref{fig:MNISTEX}(a), we observed digit 1, which clearly lay on a moon-like structure.
Therefore, Figure~\ref{fig:MK}(a) shows that 
the performance of $\mK=2, 3$, or $4$ is better than $\mK=1$ when the number of links is greater than 64.
However, in Figure~\ref{fig:MNISTEX}(b), we observe hardly any manifold structure for digits 4, 5, 6, and 7.
This observation also applies to the results in Figure~\ref{fig:MK}(b). 
The performances of $\mK = 1, 2, 3$, and $4$ are very similar to each other,
e.g., increasing the value of $\mK$ does not help.
However, 
we also notice that the increase in the number of $\mK$ does not hurt the performance of the model 
and might even enhance the performance depending on the dataset.

%

\section*{Appendix A. Mixture of Distributions}
%
\subsection*{Likelihood: Must-link Relationships}
\label{subsec:appendix:MMML}
The likelihood of the $\mcS(\cdot)$ is
\begin{align}
	\label{eq:appendix:S}
	\mcS(\mcX,\mcY,\mcZ,\mcM, 
	&
	\bvTheta) 
	:=
	p(\mcX,\mcY, \mcZ | \mcM, \bvTheta) 
	\nonumber
	=
	\prod_{m=1}^M 
	\prod_{\mk=1}^{\mK} 
	\prod_{(i,j) \in \mcM}
	\bigg
	[
	\\
	&
	\alpha_m
	\Big
	[ 
	\pi_{\mk}
	\mcN (\mbfx_i |\bmu_{\mk}, \bSigma_{\mk})	
	\Big
	]^{y_i^{\mk}}
	\Big
	[ 
	\pi_{\mk}
	\mcN (\mbfx_j |\bmu_{\mk}, \bSigma_{\mk})	
	\Big
	]^{y_j^{\mk}}
	\bigg
	]^{z_{\{ij\}}^m}
	.
\end{align}
%
%
\subsection*{Likelihood: Cannot-link Relationships}
\label{subsec:appendix:MMCL}
The likelihood of the $\mcD(\cdot)$ is
\begin{eqnarray}
	p(z_a^m,z_b^m) &=& p(z_a^m|z_b^m)p(z_b^m) 
	:= 
	p(z_b^m|z_a^m)p(z_a^m)~~ 
	\label{eqn:jointbayes}
	\\
	&=& \left\{
                \begin{array}{ll}
                  \cfrac
                  {(\alpha_m)^{z_a^m}(\alpha_m)^{z_b^m}}{1-\sum_{m'=1}^M \alpha_{m'}^2} & z_a^m \neq z_b^m\\
                  0 & z_a^m = z_b^m \label{eq:appendix:MM:propjoint}\\
                \end{array}
              \right.
              .
\end{eqnarray} 
and
\begin{align}
	\label{eq:appendix:D}
	&
	\mcD(\mcX,\mcY, \mcZ, \mcC,\bvTheta) 
	:=
	p(\mcX, \mcY, \mcZ | \mcC, \bvTheta) 
	\nonumber
	= 
	\prod_{m=1}^M
	\prod_{\mk=1}^{\mK} 
	\prod_{(a,b) \in \mcC}
	\\
	&
	\Big
	[ 
	\pi_{\mk}
	\mcN (\mbfx_a |\bmu_{\mk}, \bSigma_{\mk})	
	\Big
	]^{z_a^m~y_a^{\mk}}
	\Big
	[
	\pi_{\mk}
	\mcN (\mbfx_b |\bmu_{\mk}, \bSigma_{\mk})	
	\Big
	]^{z_b^m~y_b^{\mk}}
	p(z_a^m,z_b^m) 
	.
\end{align}
\subsection*{E-Step:}
\label{subsec:appendix:Estep}
\subsection*{Unsupervised Scenario}
\label{subsec:appendix:MM}
The expatiation $\mcL(\cdot)$ is
\begin{align}
	\label{eq:appendix:MM:Lterm:yz}
	\mbbE_{z_n^m,y_n^{\mk} | \mbfx_n}[z_n^m~y_n^{\mk}]
	&
	=
	p(z_n^m,y_n^{\mk} | \mbfx_n) 
	\\
	\nonumber
	&
	= \frac
	{\alpha_m \pi_{\mk} \mcN (\mbfx_n |\bmu_{\mk}, \bSigma_{\mk})}
	{\sum_{m'=1}^M \sum_{\mk'=1}^{\mK} \alpha_{m'} \pi_{\mk'} \mcN (\mbfx_n |\bmu_{\mk'}, \bSigma_{\mk'})}
	,
\end{align}
and
\begin{align}
	\label{eq:appendix:MM:Lterm:z}
	\mbbE_{z_n^m | \mbfx_n}[z_n^m] 
	&
	=
	p(z_n^m | \mbfx_n) 
	\nonumber
	\\
	&
	= \frac
	{\alpha_m \sum_{\mk=1}^{\mK} \pi_{\mk} \mcN (\mbfx_n |\bmu_{\mk}, \bSigma_{\mk})}
	{\sum_{m'=1}^M \sum_{\mk'=1}^{\mK} \alpha_{m'} \pi_{\mk'} \mcN (\mbfx_n |\bmu_{\mk'} \bSigma_{\mk'})}
	.
\end{align}
\subsection*{Must-link Scenario}
\label{subsec:appendix:ML}
The $\mcS(\cdot)$ is
\begin{align}
	\label{eq:appendix:MM:Sterm:zy}
	&
	\mbbE_{z^m_{\{ij\}}, y_i^{\mk} | \mbfx_i, \mbfx_j} [z^m_{\{ij\}} y_i^{\mk}]
	=
	p(z^m_{\{ij\}},y_i^{\mk}|\mbfx_i, \mbfx_j)
	\nonumber
	\\
	&
	= 
	\frac
	{
	\alpha_m
	\pi_{\mk} 
	\mcN (\mbfx_i |\bmu_{\mk}, \bSigma_{\mk})
	\sum_{\mkk=1}^{\mK}
	\pi_{\mkk}
	\mcN (\mbfx_i |\bmu_{\mkk}, \bSigma_{\mkk})
	}
	{
	Z_{\mcS}
	}
	,
\end{align}
and
\begin{align}
	\label{eq:appendix:MM:Sterm:z}
	&
	\mbbE_{z^m_{\{ij\}} | \mbfx_i, \mbfx_j}[z^m_{\{ij\}}]
	=
	p(z^m_{\{ij\}} | \mbfx_i, \mbfx_j)
	\nonumber
	\\
	&
	= 
	\frac
	{
	\alpha_m
	\sum_{\mk=1}^{\mK}
	\sum_{\mkk=1}^{\mK}
	\pi_{\mk}
	\pi_{\mkk}
	\mcN (\mbfx_i |\bmu_{\mk}, \bSigma_{\mk})
	\mcN (\mbfx_j |\bmu_{\mkk}, \bSigma_{\mkk})
	}
	{
	Z_{\mcS}
	}
	,
\end{align}
where
\begin{align}
	Z_{\mcS} = 
	\sum_{m'=1}^M 
	\sum_{\mmk=1}^{\mK}
	\sum_{\mmkk=1}^{\mK}
	\alpha_{m'}
	\pi_{\mmk}
	\pi_{\mmkk} 
	\mcN (\mbfx_i |\bmu_{\mmk}, \bSigma_{\mmk})
	\mcN (\mbfx_j |\bmu_{\mmkk}, \bSigma_{\mmkk}).
\end{align}
\subsection*{Cannot-link Scenario}
\label{subsec:appendix:ML}
The $\mcD(\cdot)$ is
\begin{align}
	\label{eq:appendix:MM:Dterm:zy}
	& 
	\mbbE_{z^m_a, y_a^{\mk} | \mbfx_a, \mbfx_b}
	[z_a^m~y_a^{\mk}]
	=
	p(z_a^m, y_a^{\mk} | \mbfx_a, \mbfx_b) 
	=
	\nonumber
	\\
	&
	\frac
	{ 
	\pi_{\mk}
	\mcN (\mbfx_a |\bmu_{\mk}, \bSigma_{\mk})
	\sum_{m'=1}^{M}
	\sum_{\mmk=1}^{\mK}
	\pi_{\mmk}
	\mcN (\mbfx_b |\bmu_{\mmk}, \bSigma_{\mmk})
	p(z_a^m,z_b^{m'})
	}
	{
	Z_{\mcD}
	}
	,
\end{align}
and
\begin{align}
	\label{eq:appendix:MM:Dterm:z}
	&
	\mbbE_{z^m_a | \mbfx_a, \mbfx_b}
	[z_a^m] 
	=
	p(z_a^m | \mbfx_a, \mbfx_b) 
	=
	\nonumber
	\\
	&
	\frac
	{ 
	\splitfrac
	{
	\sum_{\mk=1}^{\mK}
	\pi_{\mk}
	\mcN (\mbfx_a |\bmu_{\mk}, \bSigma_{\mk})
	\sum_{m'=1}^{M}
	\sum_{\mmk=1}^{\mK}
	\pi_{\mmk}
	\mcN (\mbfx_b |\bmu_{\mmk}, \bSigma_{\mmk})
	}
	{
	p(z_a^m,z_b^{m'})
	}
	}
	{
	Z_{\mcD}
	}
	,
\end{align}
where
\begin{align}
	Z_{\mcD} = 
	\sum_{m''=1}^{M}
	\sum_{m'''=1}^{M}
	\sum_{\mmmk=1}^{\mK}
	\sum_{\mmmmk=1}^{\mK}
	\pi_{\mmmk}
	\pi_{\mmmmk}
	&
	\mcN (\mbfx_a |\bmu_{\mmmk}, \bSigma_{\mmmk})
	\nonumber
	\\
	&
	\mcN (\mbfx_b |\bmu_{\mmmmk}, \bSigma_{\mmmmk})
	p(z_a^{m''},z_b^{m'''})
	.
\end{align}
%
\subsection*{M-Step}
\label{subsec:appendix:Mstep}
The mean and covariance in the $k$th cluster in the $m$th class are
\begin{align}
	\label{eq:appendix:MM:mu}
	\bmu_{\mk} 
	= 
	\frac
	{
	\sum_{n} 
	\ell_n^{\mk}
	\mbfx_n 
	+ 
	\sum_{(i,j) \in \mcM} 
	\big[
	s_i^{\mk}
	\mbfx_i 
	+ 
	s_j^{\mk}
	\mbfx_j
	\big]
	+ 
	\sum_{(a,b) \in \mcC}
	\big[
	d_{a}^{\mk}
	\mbfx_a 
	+ 
	d_{b}^{\mk}
	\mbfx_b
	\big]
	}
	{
	\sum_{n}
	\ell_n^{\mk}
	+ 
	\sum_{(i,j) \in \mcM} 
	\big[
	s_i^{\mk}
	+ 
	s_j^{\mk}
	\big] 
	+ 
	\sum_{(a,b) \in \mcC}
	\big[
	d_{a}^{\mk}
	+ 
	d_{b}^{\mk}
	\big]
	}
	,
\end{align}
\begin{align}
	\label{eq:appendix:MM:mu}
	\bSigma_{\mk} 
	= 
	\frac
	{
	\splitfrac
	{
	\sum_{n}
	\ell_n^{\mk}
	\mbfS_n^{\mk} 
	+ 
	\sum_{(i,j) \in \mcM} 
	\big[
	s_i^{\mk}
	\mbfS_i^{\mk} 
	+ 
	s_j^{\mk}
	\mbfS_j^{\mk}
	\big] 
	+ 
	\sum_{(a,b) \in \mcC}
	\big[
	}
	{
	d_{a}^{\mk}
	\mbfS_a^{\mk} 
	+ 
	d_{b}^{\mk}
	\mbfS_b^{\mk}
	\big]
	}
	}
	{
	\sum_{n} 
	\ell_n^{\mk}
	+ 
	\sum_{(i,j) \in \mcM} 
	\big[
	s_i^{\mk}
	+ 
	s_j^{\mk}
	\big] 
	+ 
	\sum_{(a,b) \in \mcC}
	\big[
	d_{a}^{\mk}
	+ 
	d_{b}^{\mk}
	\big]
	}
	,
\end{align}
where 
\begin{align}
	\label{eq:appendis:MM:shortnotation}
	\ell_n^{\mk} &= p(z_n^m, y_n^{\mk}|\mbfx_n),
	\nonumber
	\\
	s_i^{\mk} &= p(z_{\{ij\}}^m, y_i^{\mk}|\mbfx_i, \mbfx_j), 
	\nonumber
	\\
	d_{a}^{\mk} &= p(z_a^m, y_a^{\mk}|\mbfx_a, \mbfx_b),
\end{align}
and 
\begin{align}
	\mbfS_n^{\mk} = (\mbfx_n - \bmu_{\mk})(\mbfx_n - \bmu_{\mk})^T.
\end{align}
Because the mixing parameter for the cluster $\pi_{\mk}$ satisfies the summation to one,
the determination can be achieved by the Lagrange multiplier.
\begin{align}
	\label{eq:appendix:MM:pi}
	\mcQ_{\mcJ} + \lambda \bigg (\sum_{\mk=1}^{\mK} \pi_{\mk} - 1 \bigg )
\end{align}
$\lambda$ is the Lagrange multiplier. 
Taking the derivative of equation~(\ref{eq:appendix:MM:pi}) with respect to $\pi_{\mk}$,
\begin{align}
	\label{eq:appendix:MM:pi2}
	\frac
	{
	\sum_{n=1}^N 
	\ell_n^{\mk}
	+ 
	\sum_{(i,j) \in \mcM} 
	\big[
	s_i^{\mk}
	+ 
	s_j^{\mk}
	\big] 
	+ 
	\sum_{(a,b) \in \mcC}
	\big[
	d_{a}^{\mk}
	+ 
	d_{b}^{\mk}
	\big]
	}
	{
	\pi_{\mk}
	}
	+
	\lambda
	=
	0
\end{align}
By taking the derivative of equation~(\ref{eq:appendix:MM:pi}) with respect to 
$\lambda$ and equal to zero,
we then can get $\sum_{\mk=1}^{\mK} \pi_{\mk}=1$ 
and use it to eliminate the $\lambda$ in equation~(\ref{eq:appendix:MM:pi2}). 
The mixing parameter for the $k$th cluster in the $m$th mixture is given by
\begin{align}
	\label{eq:HGMM:pi3}
	\pi_{\mk}
	=
	\frac
	{
	\sum_{n=1}^N 
	\ell_n^{\mk}
	+ 
	\sum_{(i,j) \in \mcM} 
	\big[
	s_i^{\mk}
	+ 
	s_j^{\mk}
	\big] 
	+ 
	\sum_{(a,b) \in \mcC}
	\big[
	d_{a}^{\mk}
	+ 
	d_{b}^{\mk}
	\big]
	}
	{
	\sum_{\mk=1}^{\mK}
	\Big
	(
	\sum_{n=1}^N 
	\ell_n^{\mk}
	+ 
	\sum_{(i,j) \in \mcM} 
	\big[
	s_i^{\mk}
	+ 
	s_j^{\mk}
	\big] 
	+ 
	\sum_{(a,b) \in \mcC}
	\big[
	d_{a}^{\mk}
	+ 
	d_{b}^{\mk}
	\big]
	\Big
	)
	}
\end{align}
Lastly, estimating the mixing parameters for mixture $\alpha_m$ is the same as in equation~(\ref{eqn:mixing_opt}).

\bibliographystyle{spmpsci}      
\bibliography{main}   


\end{document}